\documentclass[lettersize,journal]{IEEEtran}
\usepackage{amsmath,amsfonts}
\usepackage{algorithmic}
\usepackage{algorithm}
\usepackage{array}
\usepackage{newtxtext}
\usepackage[caption=false,font=small,labelfont=rm,textfont=rm]{subfig}
\usepackage{textcomp}
\usepackage{stfloats}
\usepackage{url}
\usepackage{verbatim}
\usepackage{graphicx}
\usepackage{cite}
\usepackage{booktabs}
\usepackage{tabularx}
\usepackage{multirow}
\usepackage{multicol}
\usepackage[switch]{lineno}
\usepackage{xcolor}
\usepackage{soul}

\begin{document}

\title{WHU-Synthetic: A Synthetic Perception Dataset for 3D Multi-task Model Research}

\author{Jiahao Zhou, Chen Long, Yue Xie, Jialiang Wang, Conglang Zhang, Boheng Li, Haiping Wang, Zhe Chen, and Zhen Dong \IEEEmembership{Member, IEEE}
\thanks{This work was supported by the Major Program of National Natural Science Foundation of China (Grant No. 42394062). \textit{(Jiahao Zhou and Chen Long are co-first authors.) (Corresponding author: Zhe Chen.)}}
\thanks{Jiahao Zhou and Jialiang Wang are with the School of Remote Sensing and Information Engineering, Wuhan University, Wuhan 430079, China (e-mail: zhoujiahao@whu.edu.cn; jialiang.wang@whu.edu.cn).}
\thanks{Chen Long, Yue Xie, Conglang Zhang, Haiping Wang, Zhe Chen and Zhen Dong are with the Department of State Key Laboratory of Information Engineering in Surveying, Mapping, and Remote Sensing, Wuhan University, Wuhan 430079, China (e-mail: chenlong107@whu.edu.cn; xieyue@whu.edu.cn; zcliangyue@gmail.com; hpwang@whu.edu.cn; chenzhewhu@whu.edu.cn; dongzhenwhu@whu.edu.cn).}
\thanks{Boheng Li is with the Key Laboratory of Aerospace Information Security and Trusted Computing, Wuhan University, Wuhan 430079, China (e-mail: randy.bh.li@foxmail.com).}
}



\maketitle

\begin{abstract}
End-to-end models capable of handling multiple sub-tasks in parallel have become a new trend, thereby presenting significant challenges and opportunities for the integration of multiple tasks within the domain of 3D vision. The limitations of 3D data acquisition conditions have not only restricted the exploration of many innovative research problems but have also caused existing 3D datasets to predominantly focus on single tasks. This has resulted in a lack of systematic approaches and theoretical frameworks for 3D multi-task learning, with most efforts merely serving as auxiliary support to the primary task. In this paper, we introduce WHU-Synthetic, a large-scale 3D synthetic perception dataset designed for multi-task learning, from the initial data augmentation (upsampling and depth completion), through scene understanding (segmentation), to macro-level tasks (place recognition and 3D reconstruction). Collected in the same environmental domain, we ensure inherent alignment across sub-tasks to construct multi-task models without separate training methods. Besides, we implement several novel settings, making it possible to realize certain ideas that are difficult to achieve in real-world scenarios. This supports more adaptive and robust multi-task perception tasks, such as sampling on city-level models, providing point clouds with different densities, and simulating temporal changes. Using our dataset, we conduct several experiments to investigate mutual benefits between sub-tasks, revealing new observations, challenges, and opportunities for future research. The dataset is accessible at \url{https://github.com/WHU-USI3DV/WHU-Synthetic}.
\end{abstract}

\begin{IEEEkeywords}
Point cloud understanding, Datasets and evaluation, Multi-task learning, 3D computer vision, Environment simulation
\end{IEEEkeywords}

\section{Introduction}

\IEEEPARstart{R}{ecently}, end-to-end models have gained significant traction due to their ability to handle multiple tasks within a unified framework in autonomous driving. Notable examples such as UniAD \cite{hu2023planning} and Reasonnet \cite{shao2023reasonnet} demonstrate how integrated approaches can streamline the processing of complex, concurrent tasks. These models promise enhanced performance by learning shared representations, reducing redundancy, and improving overall efficiency. Essentially, multi-task learning forms the foundation of end-to-end models, enabling them to learn shared representations and improve overall efficiency.

\begin{figure}[t!]
    \includegraphics[width=0.98\linewidth]{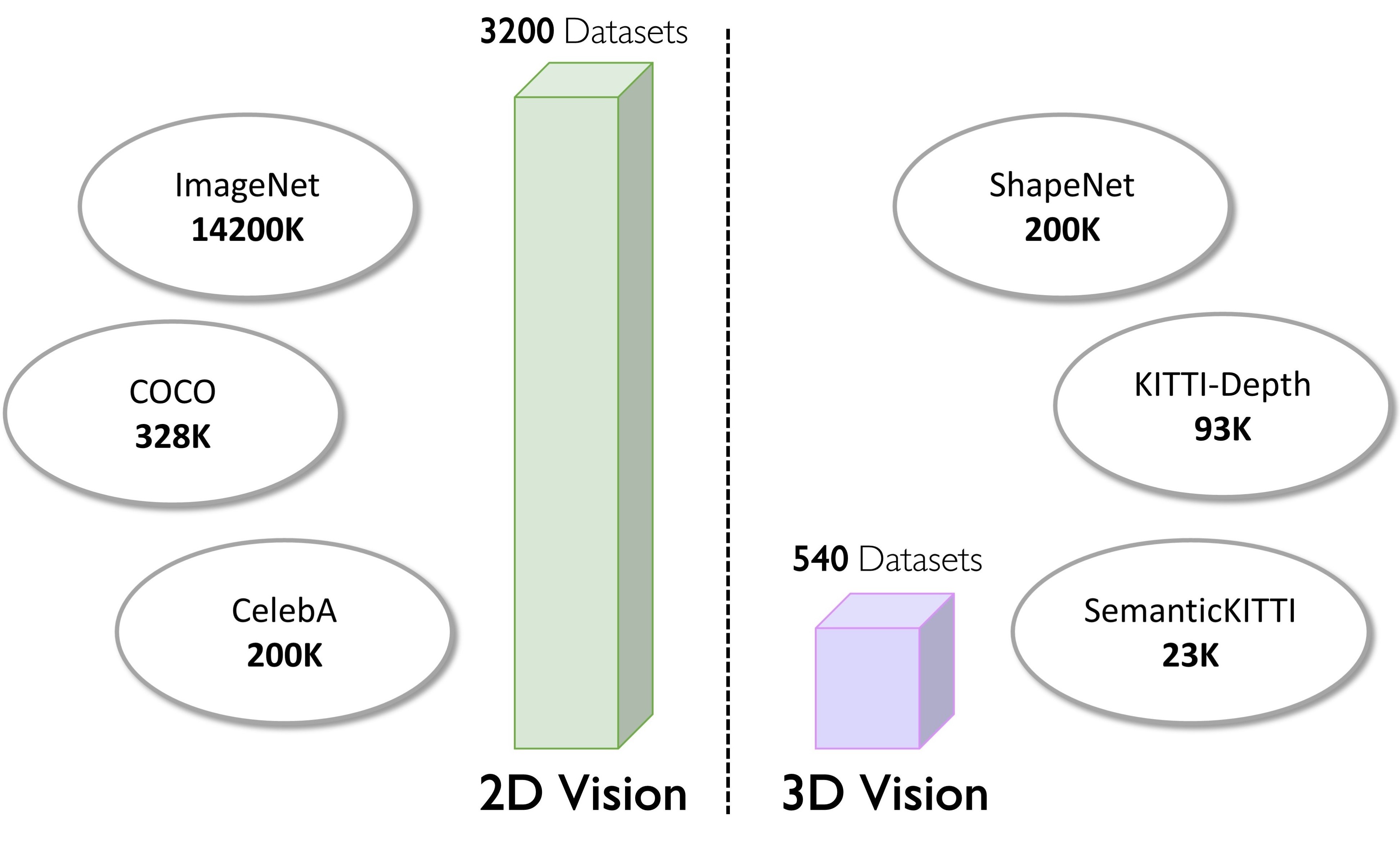}
    \caption{Comparison between dataset quantities in 2D and 3D vision, as well as the size of some classic datasets. The statistic data is from the Papers With Code \cite{paperswithcode} website.}
    \label{fig:compare}
\end{figure}

By training a model to handle several tasks concurrently, it is possible to capture commonalities and interdependencies that can enhance the performance of each task. This approach is particularly crucial for applications in intelligent perception and scene understanding, where the integration of multiple tasks is necessary for comprehensive environmental analysis \cite{li2024repvf}. 

In 2D vision, many large-scale, high-quality datasets have been designed for multi-task purposes, such as Taskonomy \cite{zamir2018taskonomy} and SA-1B \cite{kirillov2023segany}. The availability of data and relatively small domain shifts between images \cite{zhang2018overview} have facilitated the maturity of 2D multi-task models. However, in the 3D domain, such dedicated multi-task datasets are still relatively scarce. As shown in Fig. \ref{fig:compare}, 3D datasets are significantly fewer and often smaller in scale compared to their 2D counterparts \cite{deng2009imagenet, lin2014microsoft, liu2015faceattributes, Geiger2013IJRR, shapenet2015, behley2019iccv}. 

Moreover, unlike grid-based 2D images, the sparsity inherent in 3D representations results in greater domain differences between datasets compared to images, limiting the integration and use of multiple datasets. For instance, the SemanticKITTI \cite{behley2019iccv} dataset provides large-scale annotated point clouds for segmentation, but its single-density setting makes it less suitable for the upsampling task. This also leads to difficulties in domain adaptation, resulting in a decline in performance. Such a condition forces us to train multiple networks separately on different datasets, affecting the efficiency and accuracy of multi-task models and limiting the exploration of inter-task relationships \cite{vafaeikia2020brief}. Moreover, the complexity of 3D information acquisition (such as sensor calibration and error correction) makes the construction of multi-task datasets even more challenging.

\begin{table}[t]
\centering
\caption{Comparison of supported tasks of existing 3D datasets. WHU-Synthetic is the only dataset that simultaneously supports depth completion (Dep.), segmentation (Seg.), upsampling (Ups.), place recognition (Recg.), and 3D reconstruction (Recs.) in the same environmental domain.}
\label{tab:comparison}
\renewcommand\arraystretch{1.3}
\resizebox{\linewidth}{!}{
\begin{tabular}{clcccccc}
\toprule
\multicolumn{1}{l}{} & \multirow{2}{*}{\textbf{Dataset}} & \multirow{2}{*}{\textbf{Cities}} & \multicolumn{5}{c}{\textbf{Task}} \\
\cline{4-8} 
\multicolumn{1}{l}{} & &  & Dep. & Ups. & Seg. & Recg. & Recs. \\
\midrule
\multirow{6}{*}{\rotatebox{90}{\textbf{Real-world}}} & KITTI \cite{Geiger2013IJRR} & 1 & \checkmark &   &   &   & \checkmark \\
 & SemanticKITTI \cite{behley2019iccv} & 1  &   &   & \checkmark &   &   \\
 & ShapeNet \cite{shapenet2015} & object &   & \checkmark & \checkmark &   & \checkmark \\
 & Oxford RobotCar \cite{RobotCarDatasetIJRR} & 1 &   &   &   & \checkmark & \checkmark \\
 & S3DIS \cite{2017arXiv170201105A} & indoor &   &   & \checkmark &   &   \\
 & MARS \cite{li2024multiagent} & 1 &  &   &   & \checkmark & \checkmark \\
 \hline
\multirow{7}{*}{\rotatebox{90}{\textbf{Synthetic}}} 
 & SynthCity \cite{griffiths2019synthcity} & 1 &   &   & \checkmark &   &   \\
& STPLS3D \cite{Chen_2022_BMVC} & aerial &   &   & \checkmark &   &  \\
 & KITTI-CARLA \cite{deschaud2021kitticarla} & 8 & \checkmark &   & \checkmark &   & \checkmark \\
 & AIODrive \cite{Weng2020_AIODrive} & 8 & \checkmark &   & \checkmark &   &   \\
 & SHIFT \cite{shift2022} & 8 & \checkmark &   &   &   & \checkmark \\
  & AmodalSynthDrive \cite{sekkat2023amodalsynthdrive} & 8 & \checkmark &   &  \checkmark &   & \checkmark \\
  & SynRS3D \cite{song2024synrs3d} & 6 & \checkmark &   &  \checkmark &   & \checkmark \\
\cline{2-8}
 & \textbf{WHU-Synthetic (Ours)} & \textbf{11} & \textbf{\checkmark} & \textbf{\checkmark} & \textbf{\checkmark} & \textbf{\checkmark} & \textbf{\checkmark}\\
\bottomrule
\end{tabular}
}
\end{table}

A dataset collected within the same environmental domain is required for investigating the relationships among 3D sub-tasks. In this paper, we introduce WHU-Synthetic, a large synthetic dataset that aims to facilitate the exploration of multi-task models and the relationships between tasks. 
From the initial data augmentation (upsampling and depth completion), through scene understanding (segmentation), to macro-level tasks (place recognition and 3D reconstruction), we achieve this multi-tasking objective using the CARLA simulator \cite{Dosovitskiy17}. In real-world applications, annotating these tasks often consumes a significant amount of time and can introduce irreversible errors. However, we successfully utilize a complex collection setup to simultaneously gather data for multiple subtasks within the same environmental domain.

Furthermore, we implement some novel settings that are crucial for exploring advanced hypotheses to ensure a more adaptive and robust perception system, such as surface sampling on city-scale models, multi-density point clouds, and simulating temporal changes. These settings address gaps in current datasets by incorporating essential characteristics that a high-performing perception system should possess. It provides valuable data for testing the comprehensiveness of multi-task methods.

\begin{figure*}[t]
  \centering
   \includegraphics[width=\linewidth]{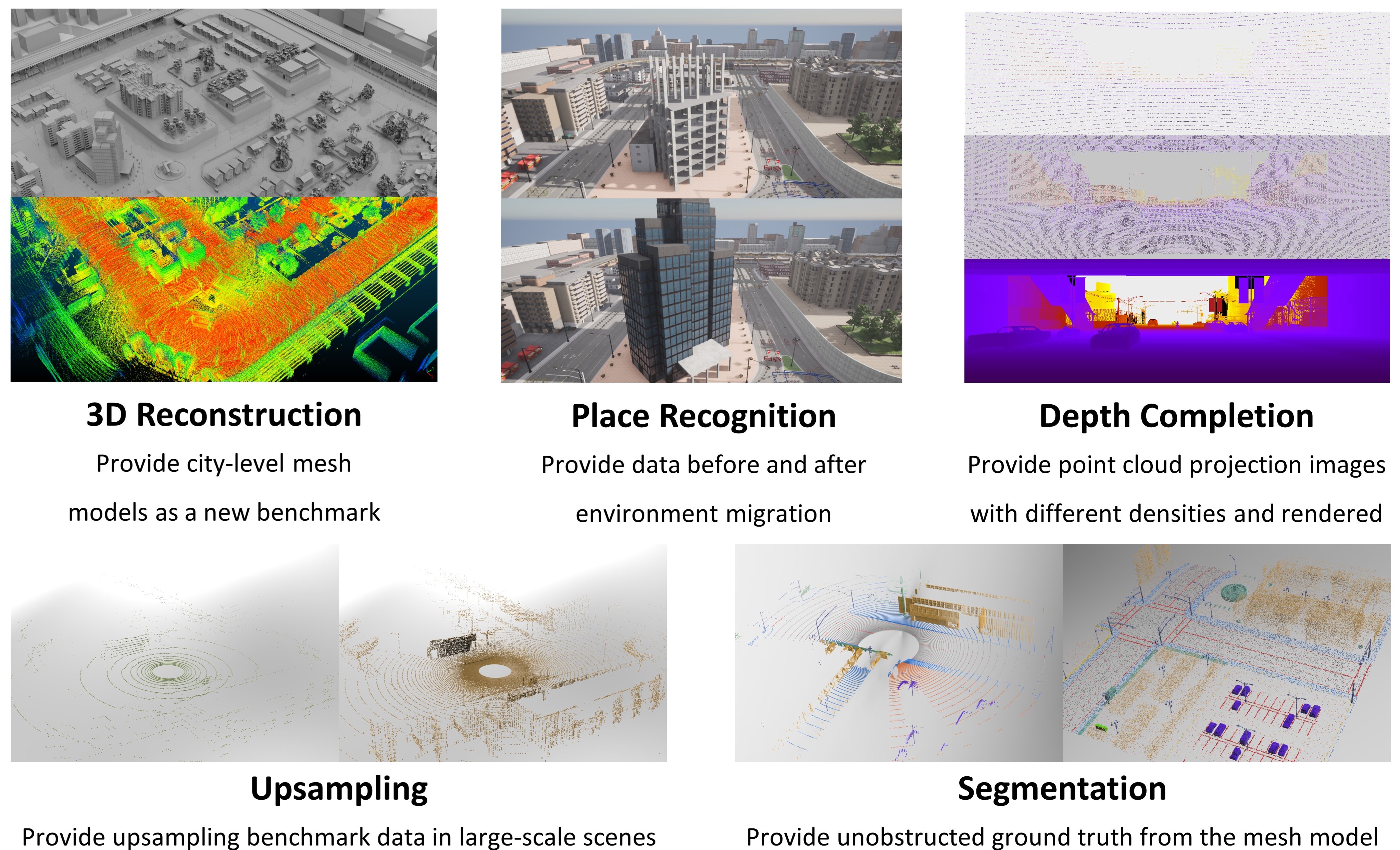}
   \caption{The distinctive features of WHU-Synthetic dataset. We implemented many novel settings, such as the stacking of sensors and the generation of annotated city mesh models. These attributes enable us to realize multi-task functionalities and data that are challenging to achieve in the real world.}
   \label{fig:Summary}
\end{figure*}

After constructing the dataset, we test the performance of each subtask using mainstream methods to create baselines. Next, to investigate the dataset's ability to explore the relationships between multiple tasks, we design a multi-task framework to leverage the features between different sub-tasks, thereby achieving semantic segmentation and upsampling concurrently. Additionally, We also test several multi-task networks on the dataset to explore the auxiliary relationships between specific tasks. The experimental results tentatively reveal the potential of the proposed dataset in uncovering the mutually beneficial relationships among sub-tasks. Beyond multi-task model development, we believe our dataset can assist researchers in exploring unified models, fostering innovation, and progressively integrating target tasks within 3D perception (See Fig. \ref{fig:Summary}).

The main contributions of our work are summarized as follows:
\begin{itemize}
    \item We introduce the WHU-Synthetic dataset, the first synthetic dataset that allows the validation of multiple tasks within the same environmental domain.
    \item The data collected in virtual environments yields novel attributes that are difficult to acquire in real-world situations, thereby providing new challenging benchmarks.
    \item We design a series of experiments to explore the potential of the dataset, including a multi-task framework with a shared feature extractor, as well as methods that use auxiliary tasks to support the main task. The experimental results highlight the dataset's versatility and capability to advance future research.
\end{itemize}

\section{Related Works}
Existing 3D datasets often exhibit limitations by being restricted to a single functionality. Tab. \ref{tab:comparison} shows the functionalities provided by mainstream datasets. Next, we introduce the dataset for 3D multi-task learning and review the current state of research.

\subsection{Single-task Dataset}
\noindent \textbf{Depth Completion. }     
Currently, datasets for depth completion tasks are relatively scarce. KITTI DC \cite{Uhrig2017THREEDV} is a widely used large-scale outdoor dataset that contains over 93,000 semi-dense depth maps with the corresponding raw sparse LiDAR scans and RGB images \cite{hu2022deep, long2024sparsedc, yan2024uncertainty}. Due to the presence of noise, this can affect the training of the model, which in turn can lead to errors in the actual results. The DenseLivox \cite{yu2021grayscale} dataset is collected with much denser depth maps than KITTI. DeepLiDAR \cite{Qiu_2019_CVPR} provides a synthetic dataset generated by the CARLA simulator. In comparison, our dataset provides accurate ground truth depth obtained through rendering and images projected from point clouds with varying densities in the same frame.

\noindent \textbf{Upsampling. } 
There are various point cloud datasets \cite {wu20153d, shapenet2015, uy-scanobjectnn-iccv19, bogo2014faust} used to evaluate the performance of point cloud upsampling algorithms. For object-level upsampling tasks, researchers usually downsample the ground truth as the input. Some researchers choose to build their dataset for training and testing of point cloud upsampling models \cite{yu2018pu, Qian_2021_CVPR, yu2018ec, li2019pu, qian2020pugeo, wu2023omniobject3d}. Compared to ours, these datasets do not provide data collected in large-scale outdoor scenes.

\noindent \textbf{Segmentation. }  
For real-world datasets, SemanticKITTI \cite{behley2019iccv} provides point-wise annotated point clouds. Recently, several similar vehicle-mounted outdoor scene datasets \cite{roynard2017parislille3d, serna2014paris, vallet2015terramobilita, nuscenes} with richer classes have been released. Besides, large-scale point clouds can be obtained through airborne LiDAR scanning \cite{9469802, ye2020lasdu, zolanvari2019dublincity} or reconstruction by photogrammetry \cite{Li2020Campus3D, hu2020towards, can2021semantic}. For synthetic datasets, SynthCity \cite{griffiths2019synthcity} is the pioneering synthetic point cloud dataset that utilizes the Blensor plugin in Blender to simulate LiDAR scanning. Paris-CARLA-3D \cite{deschaud2021paris} consists of real-world point clouds and synthesized point clouds, both annotated with the same categories, enabling exploration of domain transfer. SynLiDAR \cite{xiao2022transfer} is developed using the UE4 engine for the same motivation. STPLS3D \cite{Chen_2022_BMVC} employs a 3D scene generation engine to synthesize virtual scenes, using a UAV simulator for photogrammetry and generating labeled point clouds. Our dataset simultaneously provides annotated point clouds obtained from LiDAR scanning and the city mesh model.

\noindent \textbf{Place Recognition. } 
There are currently a limited number of large-scale place recognition datasets available. Oxford RobotCar \cite{RobotCarDatasetIJRR} and Mulran \cite{gskim-2020-mulran} are created using a LiDAR sensor mounted on a car that repeatedly drives through each region at different times. Our dataset simulates environmental migration to more effectively explore the robustness of place recognition methods.

\noindent \textbf{3D Reconstruction. }
Many scene-level datasets have been used for 3D reconstruction. LiDAR-based SLAM is a widely adopted method. KITTI Odometry \cite{Geiger2013IJRR} is now the most popular benchmark for odometry evaluation. The NCLT dataset \cite{ncarlevaris-2015a} comprises 27 extended sequences captured by Segway platforms. The UrbanNav dataset \cite{hsu2021urbannav} offers a challenging data source for advancing the study of accurate and robust positioning in urban canyons. Our dataset includes city mesh models, enabling the development of novel evaluation methods.

While the aforementioned datasets have achieved commendable outcomes within their respective domains, it is imperative to acknowledge the profound discrepancies among 3D sensors (including frequency, channels, and the number of emission points), which lead to significant domain shifts among these datasets. This necessitates individual training on each dataset, further complicating the integration of networks for various sub-tasks. There is a pressing need for a unified dataset that would enable researchers to design a multi-task framework in a singular effort.

\subsection{3D Multi-task Dataset}
Recently, some synthetic datasets with multiple functionalities have been released. KITTI-CARLA \cite{deschaud2021kitticarla} utilizes sensors identical to the KITTI dataset within a virtual environment. AIODrive \cite{Weng2020_AIODrive} is generated with a specific focus on high-density and long-range LiDAR. SHIFT \cite{shift2022} features comprehensive annotations, enabling the investigation of performance degradation as domain shift increases. 
AmodalSynthDrive \cite{sekkat2023amodalsynthdrive} provides multi-view camera images, 3D bounding boxes, LiDAR data, and odometry for 150 driving sequences. However, it still focuses on multiple amodal perceptions of 2D images. SEVD \cite{aliminati2024sevd} is a first-of-its-kind multi-view ego, and fixed perception synthetic event-based dataset using multiple dynamic vision sensors. Open MARS dataset \cite{li2024multiagent} significantly advances multi-task learning for autonomous driving by providing a comprehensive, multi-agent, multi-traversal, and multimodal dataset that fosters research in collaborative perception, and 3D reconstruction. SynRS3D \cite{song2024synrs3d} is a synthetic remote sensing dataset that contributes to global 3D semantic understanding by providing diverse and high-resolution imagery with precise height information and land cover types.

Currently, most of these datasets are collected using the CARLA simulator, but they primarily address multi-task requirements by providing annotations on 2D images. As shown in Tab. \ref{tab:comparison}, there are no datasets specifically designed to meet the needs of 3D downstream sub-tasks. In 3D vision, sub-tasks are delineated with greater specificity and cannot be simply represented through images with different annotations. Our dataset successfully provides the necessary data for a variety of 3D vision sub-tasks within the same environmental domain. Moreover, we have collected many forms of data that are challenging to obtain in the real world, such as scene-level point cloud upsampling, offering more solid support for 3D perception tasks.

\subsection{3D Multi-task Learning}
Multi-task learning has become an important topic in machine learning. Previous approaches have attempted to strike this balance by leveraging gradient-based learning of task affinities in encoded representations, employing task-conditioned gates in the decoder, attention-based task similarities \cite{misra2016cross, vandenhende2020mti, xu2018pad}, or weighted task losses \cite{kendall2018multi, chen2018gradnorm, lu2019multi, zhang2013multi, yan2024dynamic}. In multi-task learning based on 2D images, researchers output images with different annotations to accomplish multiple downstream sub-tasks. Different images can simply share a common base image.

However, in 3D computer vision, the practice of multi-task learning is relatively sparse. Several works \cite{hassani2019unsupervised} rely on heavily designed multiple task heads \cite{zhao2024robust} and the combination of various loss functions \cite{wang2024multi, yan2024dnact, yuan2023prototype}. For example, Hassani et al. \cite{hassani2019unsupervised} propose an unsupervised multi-task model for jointly learning point and shape features in point clouds. To extract highly effective features, GPA-Net \cite{shan2023gpa}, comprising an encoder and multi-task decoders, proposes a novel graph convolution kernel termed GPAConv. Liu et al. \cite{liu2024point} integrate several tasks into a unified input-output space, enabling training across various tasks within the same framework without task-specific head designs.

In recent years, an increasing number of researchers have begun to focus on the mutually beneficial relationships between downstream point cloud tasks and the auxiliary roles of ancillary tasks. Ye et al. \cite{ye2023lidarmultinet} presents a unified multi-task LiDAR-based network that achieves state-of-the-art performance across 3D object detection, semantic segmentation, and panoptic segmentation by integrating these tasks into a single, efficient, and versatile framework. LiSD \cite{xu2024lisd} is a framework that synergistically enhances lidar-based segmentation and detection through an efficient voxel encoder-decoder approach, leading to top results on nuScenes. PAttFormer \cite{lang2024point} is a novel point-based architecture for efficient joint semantic segmentation and object detection in LiDAR point clouds, which achieves competitive performance with significantly reduced network size and improved computational speed. RepVF \cite{li2024repvf} is a unified vector field representation for multi-task 3D perception in autonomous driving, that enhances efficiency and effectiveness by reducing computational redundancy and feature competition.

Some researchers have also explored the beneficial impact of auxiliary tasks on primary tasks. PatchAugNet \cite{zou2023patchaugnet} uses patch reconstruction to aid in feature extraction, thereby achieving higher place recognition accuracy. PADLoC \cite{arce2022padloc} uses panoptic information to create a new branch loss, thereby improving the effectiveness of loop closure detection and registration. However, due to the lack of 3D multi-task datasets, these methods can only utilize existing datasets to explore limited task relationships. Most of the research relies on semantic annotation to assist in finding multi-task relationships, which hinders deeper and more comprehensive multi-task learning.

\subsection{Synthetic Data Generation}
3D perception simulation technology has become increasingly sophisticated, enabling the support of realistic image and LiDAR scan simulations to reduce domain shifts between simulated and real-world datasets.

Many simulators, such as Sim4CV \cite{Müller_Casser_Lahoud_Smith_Ghanem_2018}, Nvidia Drive \cite{bojarski2016end} and Matlab Automated Driving Toolbox \cite{AutomatedDrivingToolbox}, can be utilized for synthetic data generation. However, most of these simulators are not open-source and lack a free-to-use license, making modifications difficult and disallowing derivative products. Among the open-sourced simulators, AirSim \cite{airsim2017fsr} and CARLA Simulator \cite{Dosovitskiy17} are popular choices due to their comprehensive documentation and diverse sensors. While AirSim has advantages in aerial data capture, it does not offer the same level of low-level control over agents as CARLA does. Additionally, commercial video games like GTAV \cite{Richter_Hayder_Koltun_2017} can be used for synthetic data generation, but they cannot control scene elements at a low level. Therefore, we have opted to use CARLA for data generation, as it provides the highest degree of flexibility and customization.

\section{WHU-Synthetic Dataset}
\begin{figure}[!th]
    \centering
    \includegraphics[width=\linewidth]{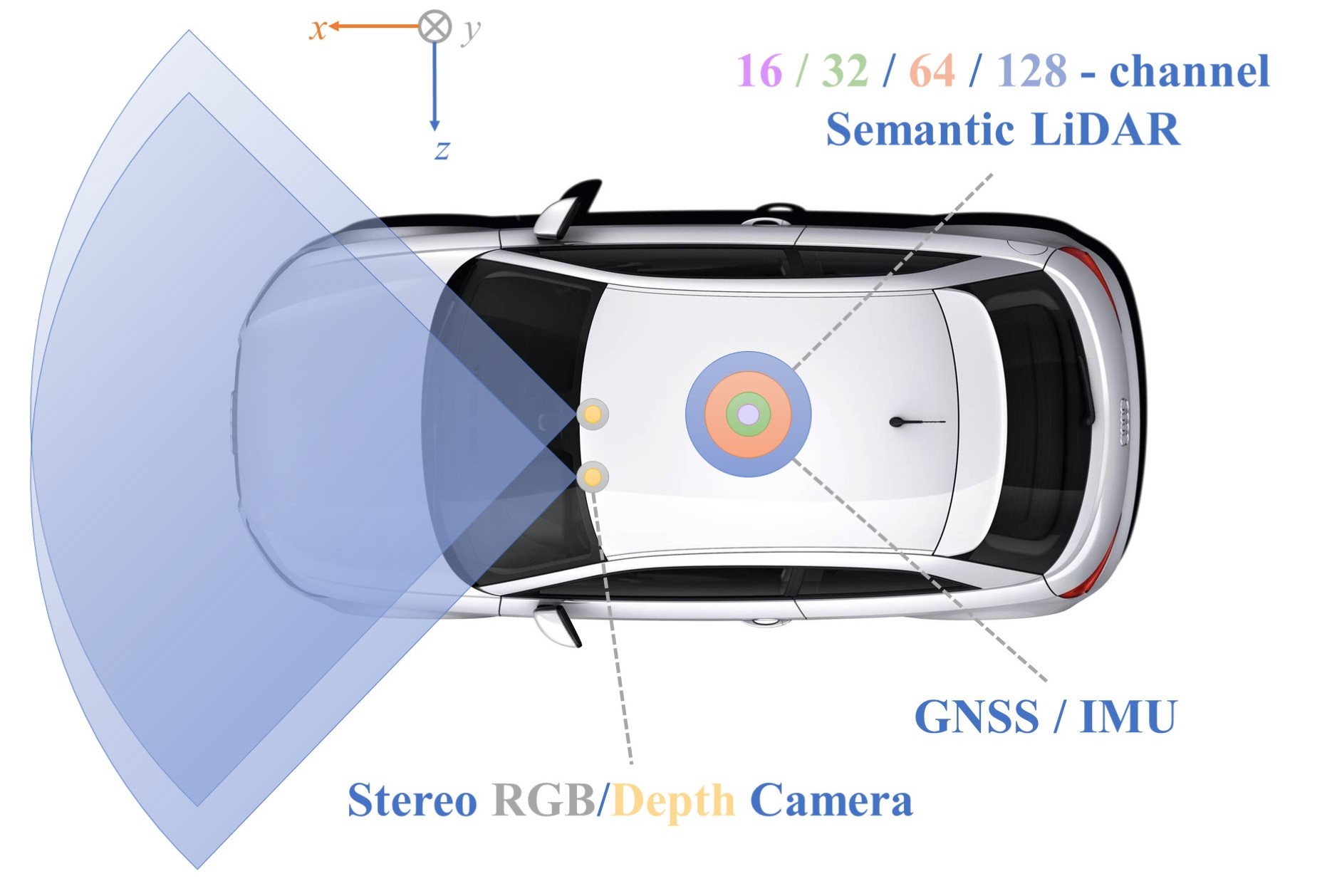}
    \caption{The vehicle system and the sensor layout. We successfully stack LiDARs with different channel counts at the same location to simulate point clouds of different densities at the sensor level, and the data are automatically annotated. The point clouds generated by LiDAR sensors share the same coordinate system and do not produce mutual occlusions.}
    \label{fig:sensor}
\end{figure}

In order to make our dataset available for different sub-tasks in the same frame, we design a data acquisition plan capable of accommodating multiple sub-tasks. Specific data details are presented in Tab. \ref{tab:briefintro}, the sensor setup is illustrated in Fig. \ref{fig:sensor}, and the parameters of the sensors are listed in Tab. \ref{tab:sensor}. Different sub-tasks can share the same frame. WHU-Synthetic has more than 140,000 fully annotated frames, occupies more than 2,500 GB, and has a total of 11 scenes ranging from urban to rural, covering an area of over 23,000 hectares. We additionally provide .OBJ city mesh models and semantic-annotated point clouds sampled from their surfaces.

\begin{table*}[th]
\centering
\caption{In the WHU-Synthetic dataset, a single frame contains data that can be used to train multiple tasks simultaneously. To make it easier to work with, we outline the specific data items that may be relevant to individual sub-tasks in the table. However, it's important to emphasize that all of this data is captured at the same moment, which is key to our multi-task design. Additionally, for the sub-tasks of 3D reconstruction and place recognition, which require fixed routes, we have designed a series of specific routes and collected data separately.}
\label{tab:briefintro}
\renewcommand\arraystretch{1.5}
\begin{tabularx}{\textwidth}{llX}
\toprule
\textbf{Function} & \textbf{Format} & \textbf{Brief Description} \\
\midrule
Depth Completion & KITTI DC \cite{Uhrig2017THREEDV} & 
16/32/64/128-channel LiDAR projected image; monocular RGB image;  rendered depth ground truth.\\

Upsampling & - &
16/32/64/128-channel LiDAR scan.\\

Segmentation & SemanticKITTI \cite{behley2019iccv} & 
16/32/64/128-channel LiDAR scan with semantic \& instance label. \newline Provide semantic-annotated point clouds sampled from the city mesh model.\\

Place Recognition & Oxford RobotCar \cite{RobotCarDatasetIJRR} & 
Monocular RGB image; rendered depth image; 64-channel LiDAR scan with pose and location. \newline Four partially overlapping routes collected before and after environmental migration.\\

3D Reconstruction & KITTI Odometry \cite{Geiger2013IJRR} &
Stereo RGB image; 64-channel LiDAR scan with pose and location. \newline Provide city mesh models as ground truth.\\
\bottomrule
\end{tabularx}
\end{table*}

\begin{table}[!t]
\centering
\caption{Sensor parameter settings. We meticulously configure our setup by the parameters of KITTI\cite{Geiger2013IJRR} sensors and added noise and post-effects to ensure the realism of the data, aiming to minimize domain shift. }
\label{tab:sensor}
\renewcommand\arraystretch{1.3}
\begin{tabularx}{\linewidth}{lX}
\toprule
\textbf{Sensor} & \textbf{Parameters} \\
\midrule
LiDAR & 16/32/64/128-channel, $360^\circ$ horizontal FoV, $-13.4^\circ$ to $13.4^\circ$ vertical FoV, 10Hz frequency, $\leq$ 120m range. With added noise.\\
RGB Camera & Resolution of 1216 × 352, FoV of $90^\circ$, Aperture of $f/1.4$. With post-process effects.\\
Depth Camera & Same as RGB Camera.\\
GNSS/IMU & 10Hz frequency.\\
\bottomrule
\end{tabularx}
\end{table}

We employ LiDAR with different channel counts, RGB cameras, Depth cameras, and GNSS/IMU for data collection purposes. To enhance realism and reduce the domain gap, we added noise into the point clouds. Points are randomly perturbed along the laser ray's direction, introducing noisy distance measurements. Taking the feature that virtual environments can be freely edited, we implement the stacking of LiDAR sensors at the same location, generating point clouds that do not obscure each other and share the same coordinate system. Thus we realized the collection of point clouds with different densities at the sensor level, providing various configurations for the dataset. We use a randomized approach to divide the training set, validation set and test set with a ratio of 8:1:1.

\begin{figure*}[hp]
\centering
    \subfloat[Town01]{\includegraphics[width=0.24\linewidth]{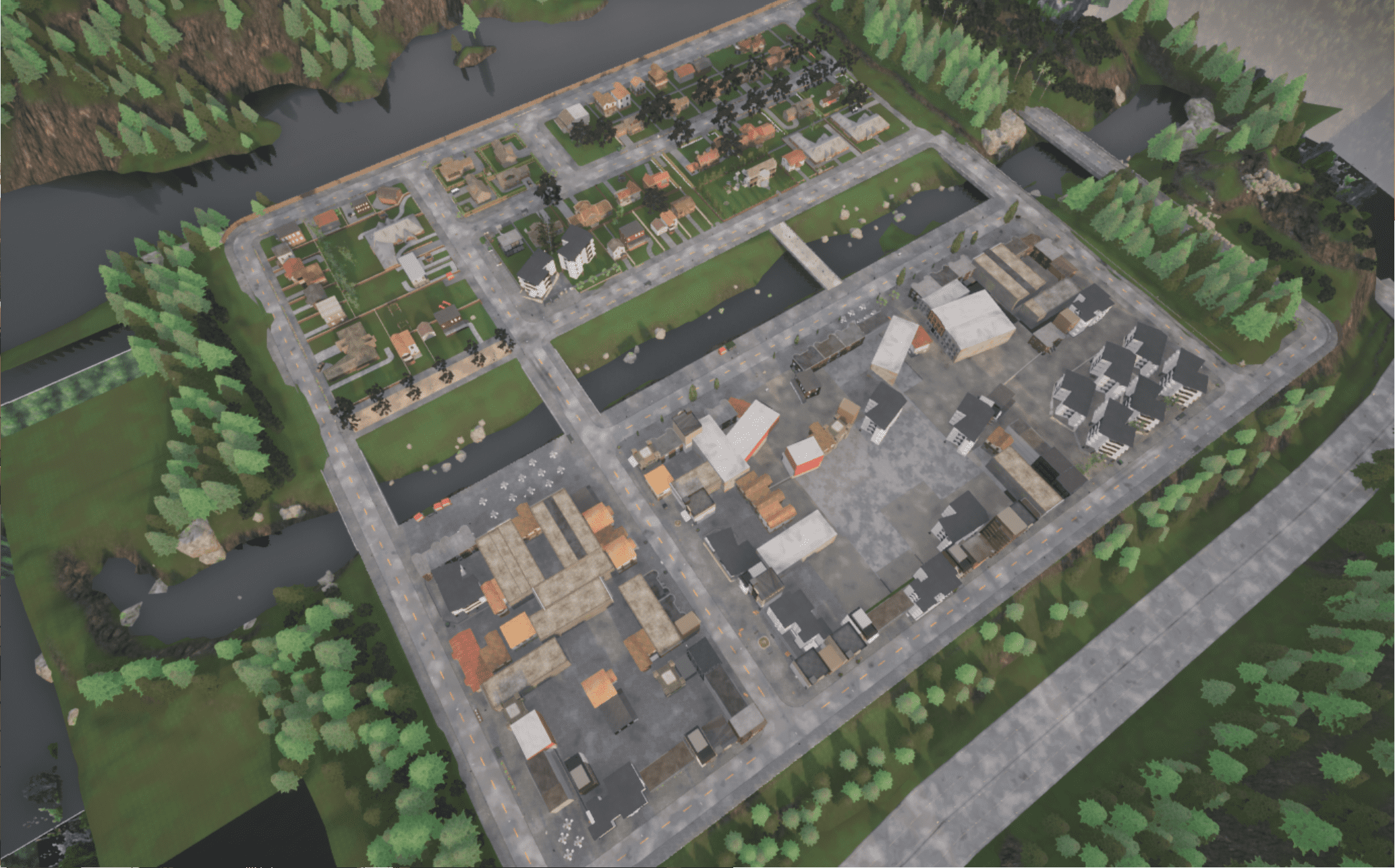}}
    \hfil
    \subfloat[Town02]{\includegraphics[width=0.24\linewidth]{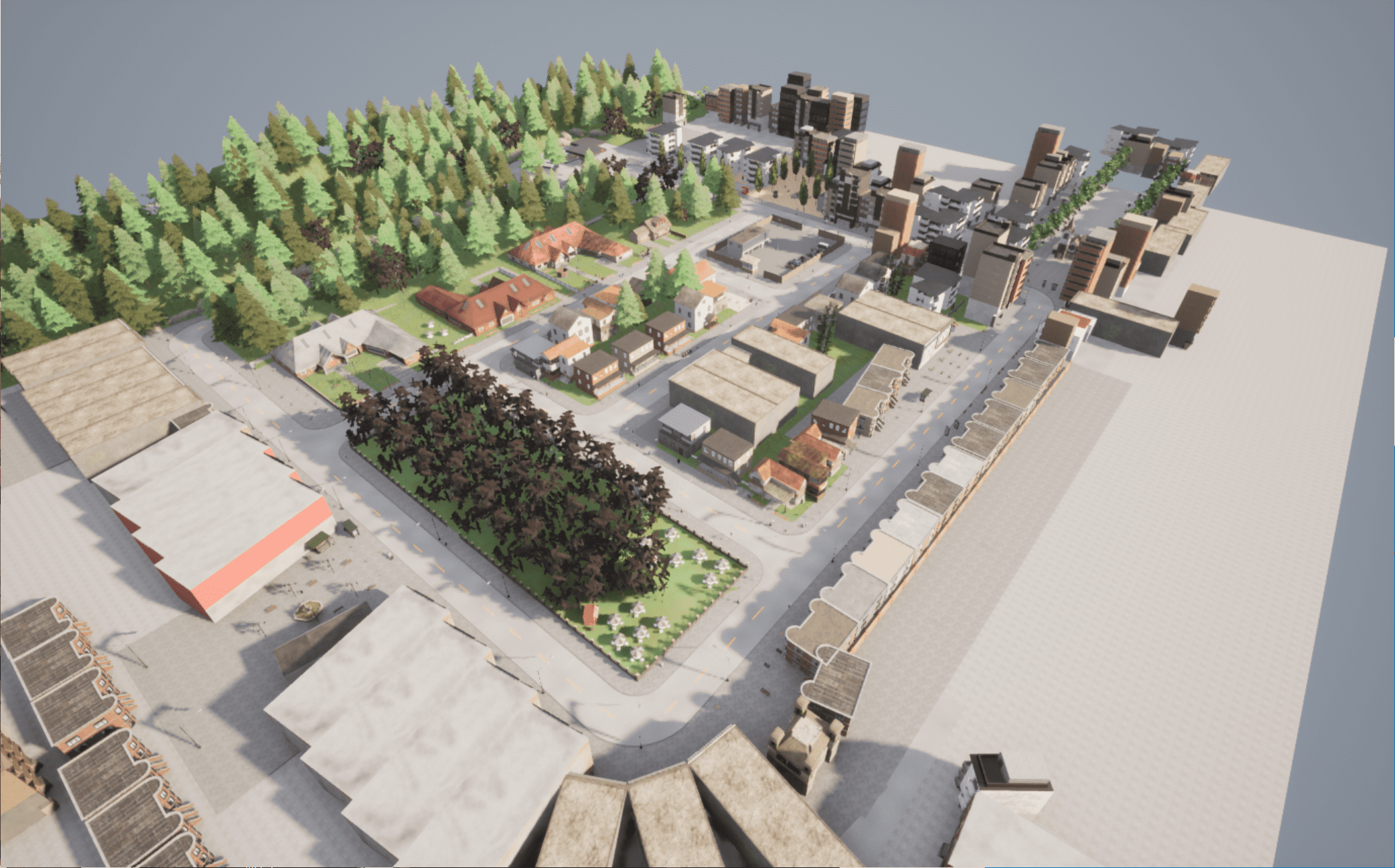}}
    \hfil
    \subfloat[Town03]{\includegraphics[width=0.24\linewidth]{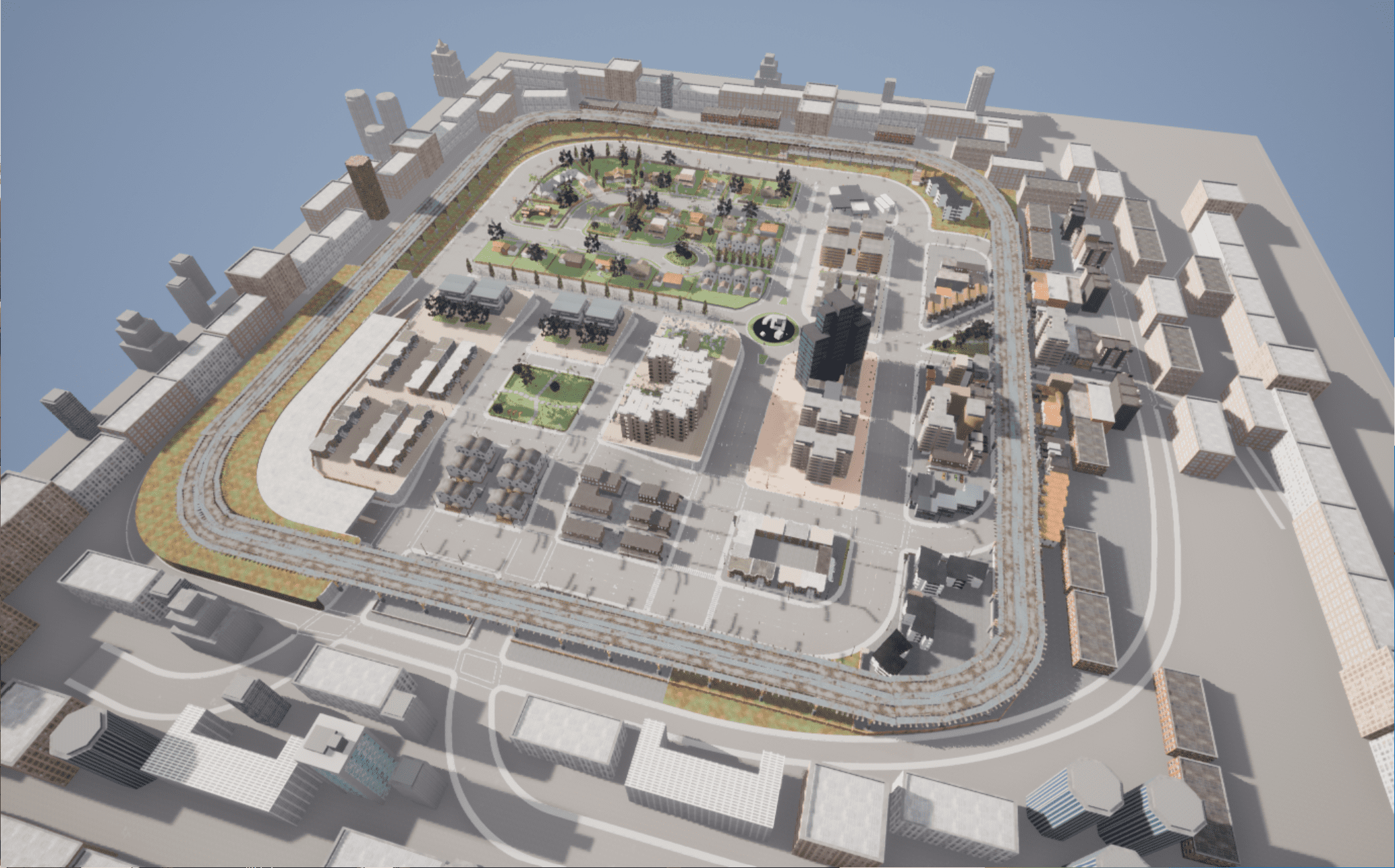}}
    \hfil
    \subfloat[Town04]{\includegraphics[width=0.24\linewidth]{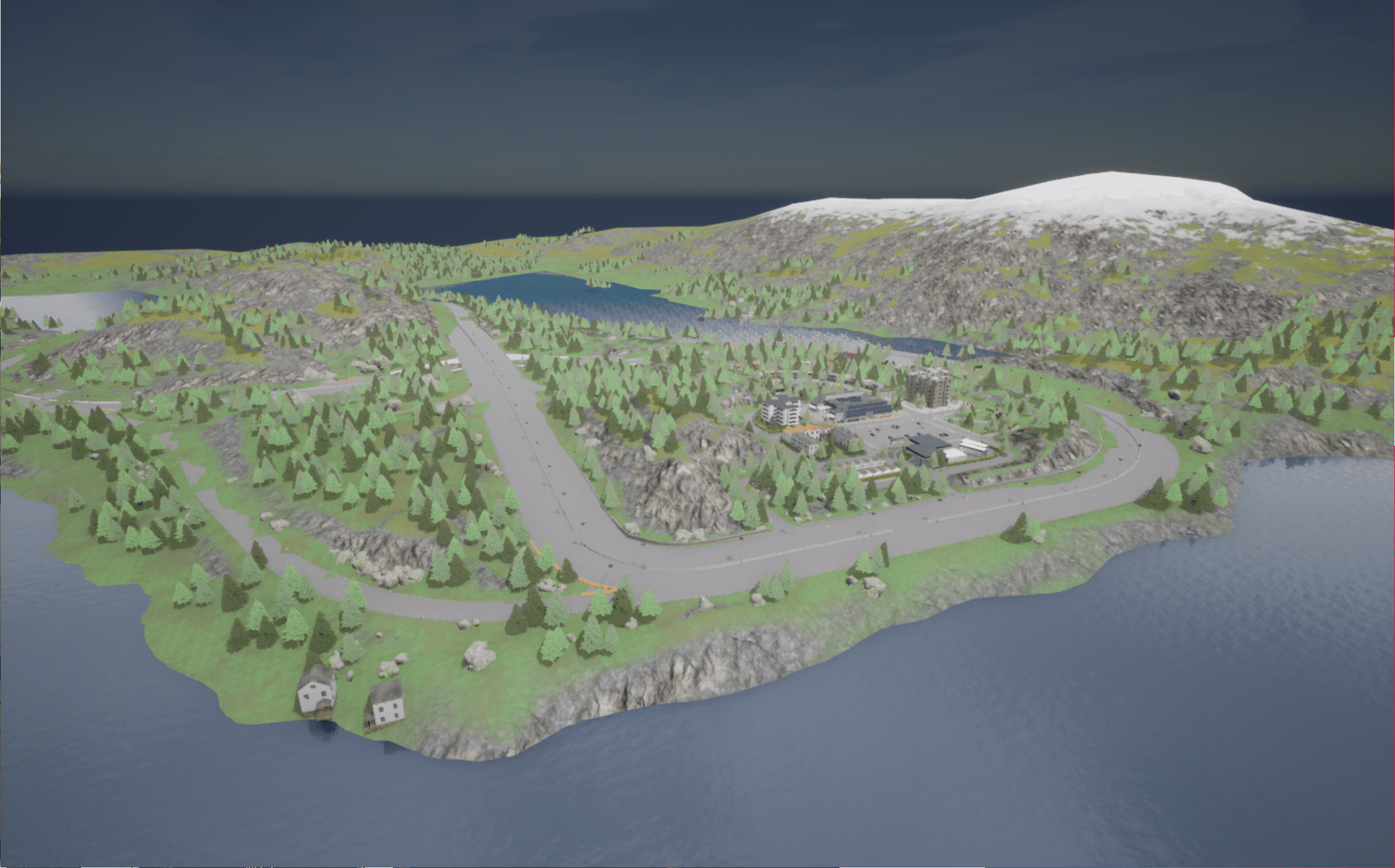}}
    \\
    \subfloat[Town05]{\includegraphics[width=0.24\linewidth]{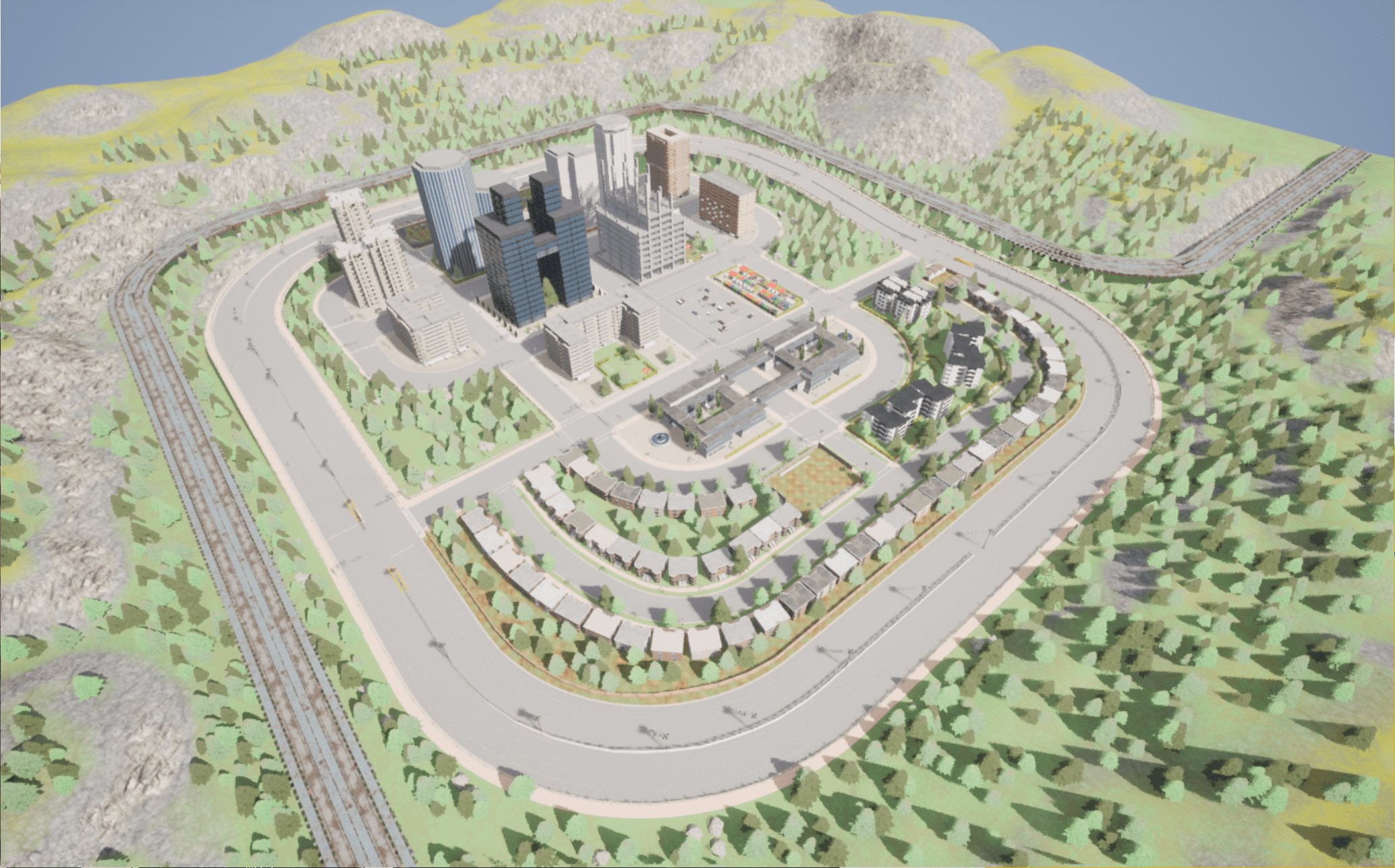}}
    \hfil
    \subfloat[Town06]{\includegraphics[width=0.24\linewidth]{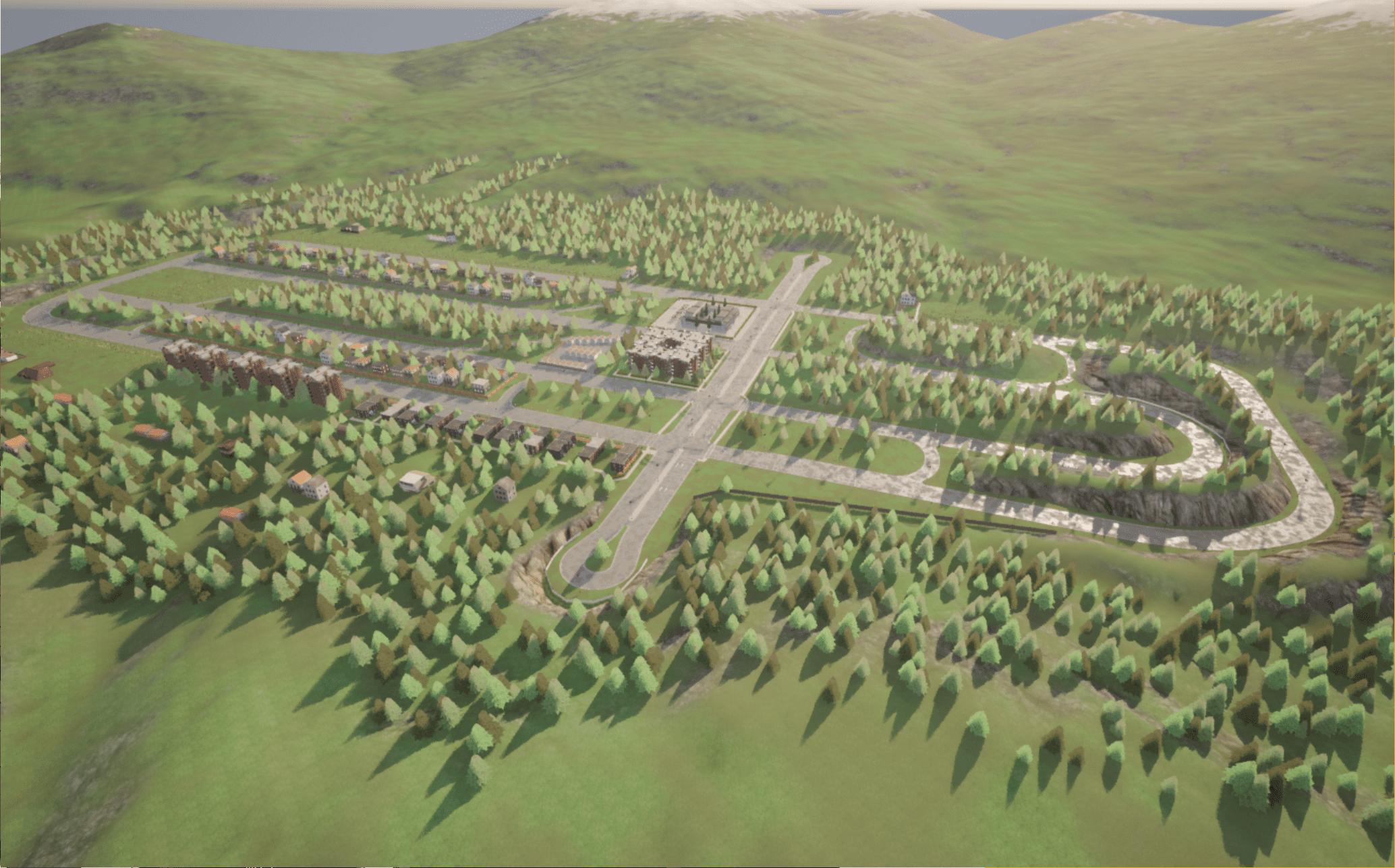}}
    \hfil
    \subfloat[Town07]{\includegraphics[width=0.24\linewidth]{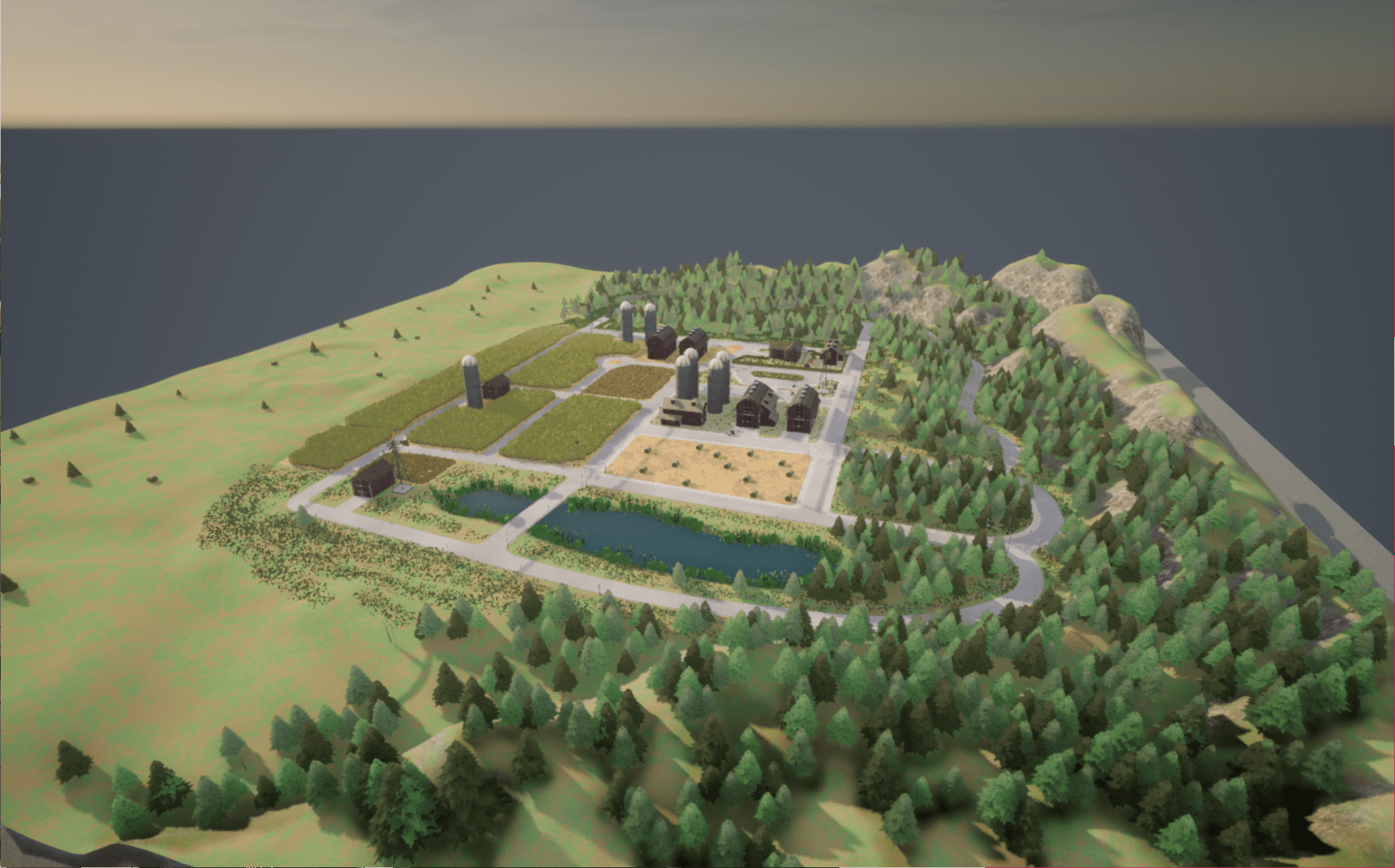}}
    \hfil
    \subfloat[Town10]{\includegraphics[width=0.24\linewidth]{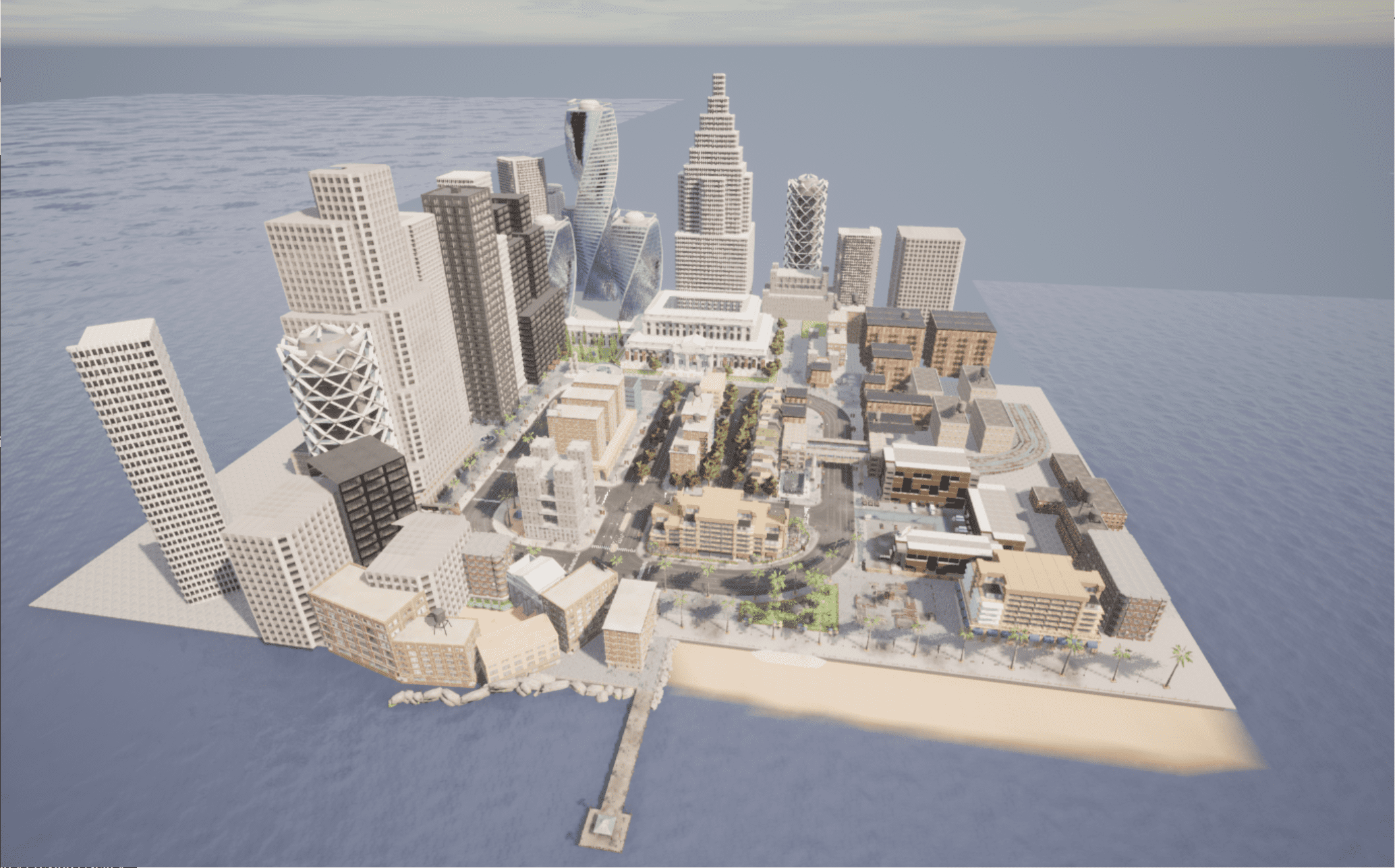}}
    \\
    \centering
    \subfloat[Town12]{\includegraphics[width=0.32\linewidth]{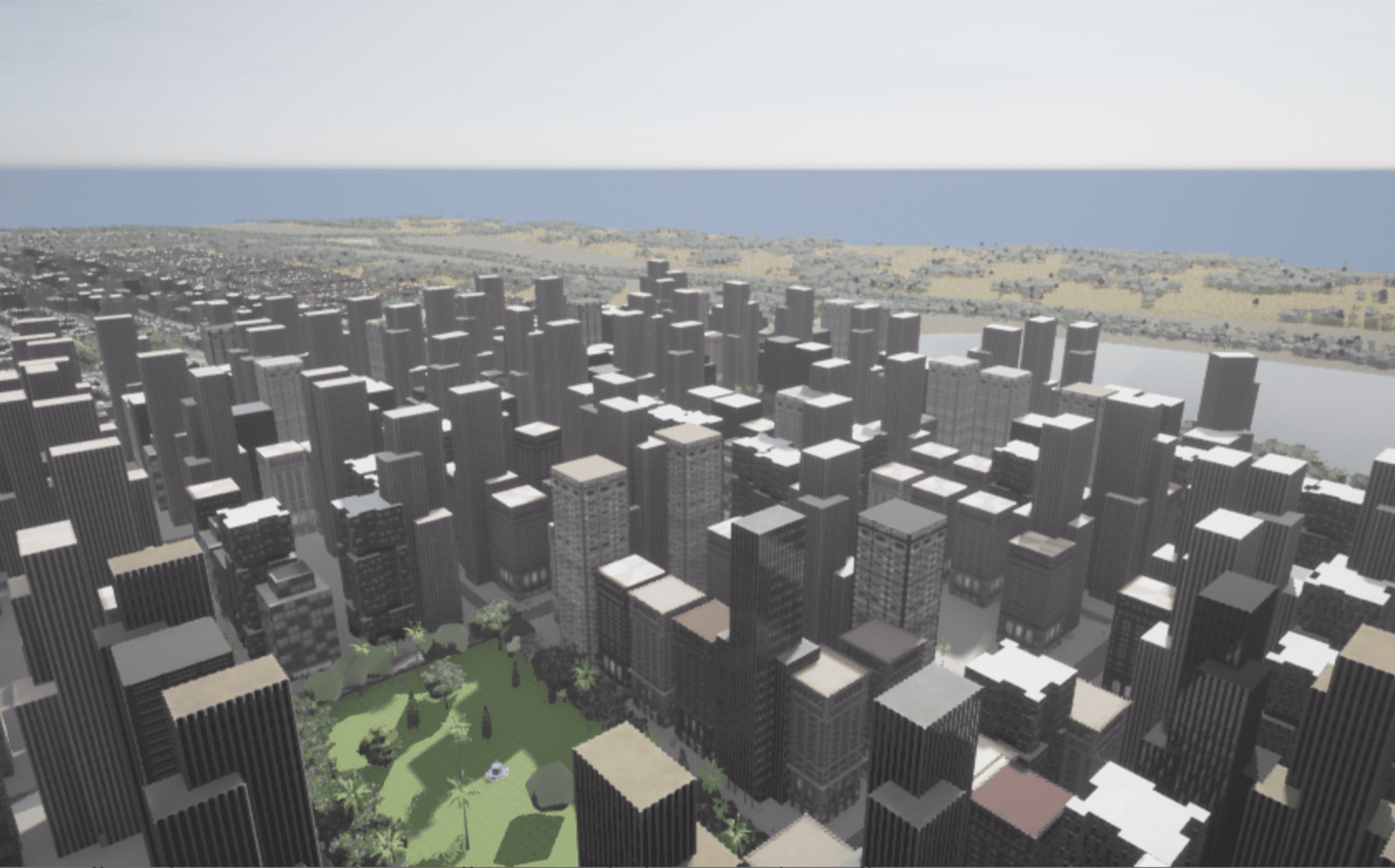}}
    \hfil
    \subfloat[Town13]{\includegraphics[width=0.32\linewidth]{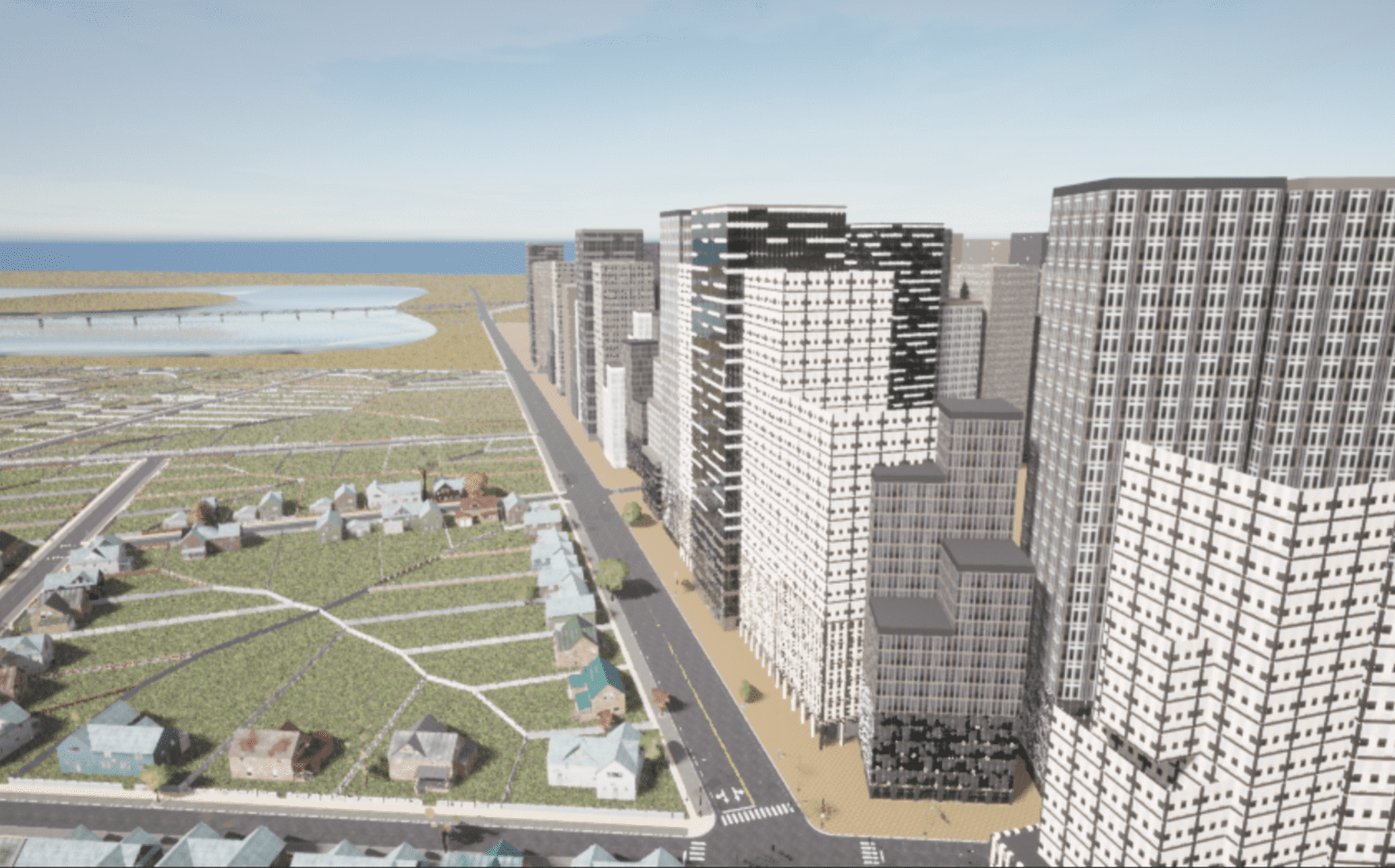}}
    \hfil
    \subfloat[Town15]{\includegraphics[width=0.32\linewidth]{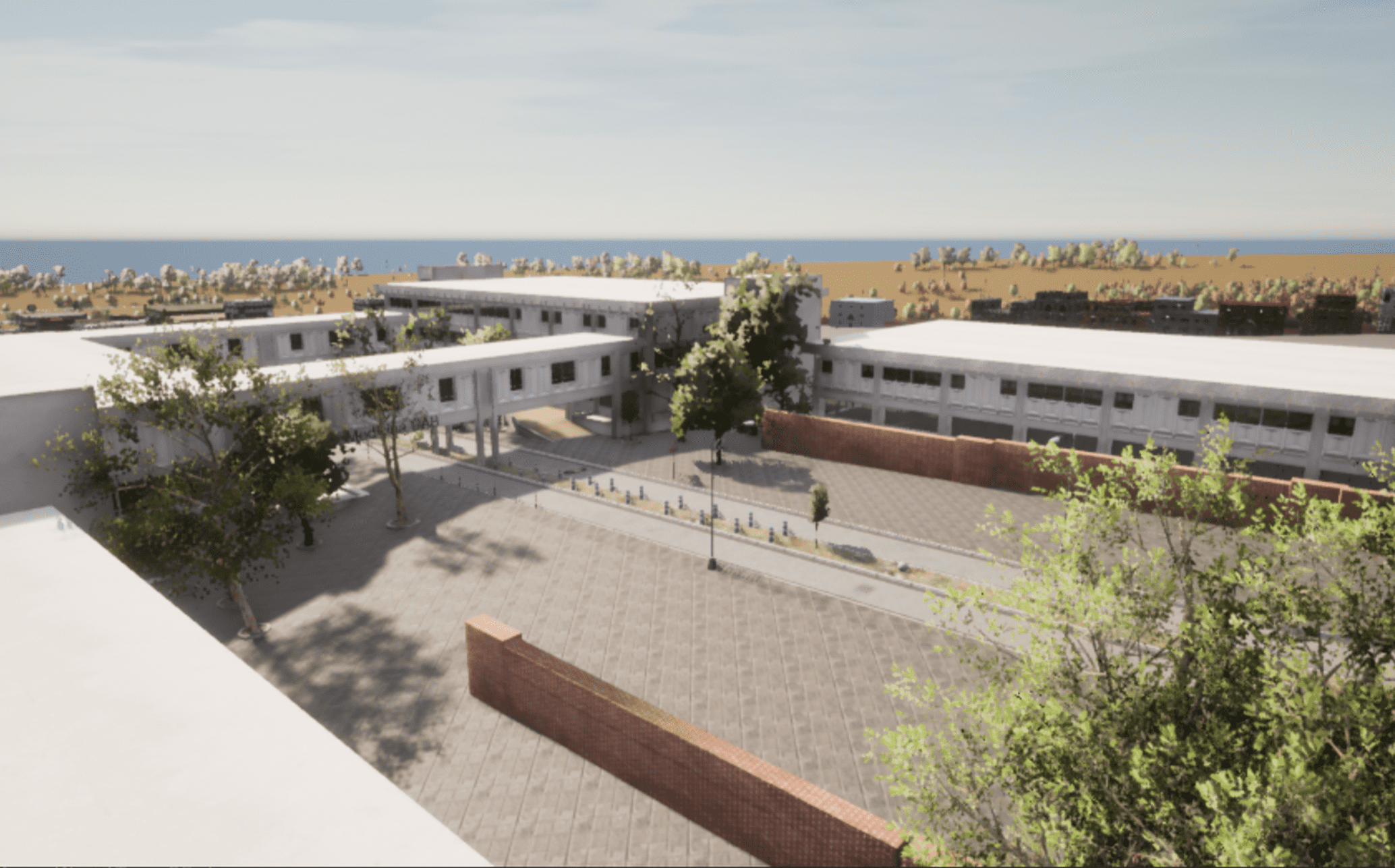}}
\caption{The 11 city maps used during the collection process. }
\label{fig:map}
\end{figure*}

We allow the vehicle to follow random routes in the city map and simulate real traffic conditions to collect data in different scenarios. These city maps vary in size, with the largest covering 10,000 hectares and encompassing a range of scenes from urban to rural, and from streets to highways. Weather conditions also vary on each map, from sunny to rainy to foggy, from dawn to noon, dusk to night. The scene thumbnails can be found in Fig. \ref{fig:map}. Note that for sub-tasks such as 3D reconstruction and place recognition, which require fixed routes, we design several separate routes that can fulfill the task requirements. These routes overlap with the previously random sections, and the frames on these fixed routes can also meet the requirements of the remaining three sub-tasks. 

In environments, annotations for sub-tasks can be automatically performed based on identical data, which is the main reason why our dataset can help multi-task learning. We provide instance and semantic segmentation labels on 28 classes and the real depth map generated by the renderer. We also obtain semantic-labeled point clouds through Monte Carlo sampling \cite{hastings1970monte} from the city model, enabling new benchmarks for scene-level completion and 3D reconstruction tasks.

To facilitate the simultaneous learning of multiple subtasks and the exploration of relationships between them, WHU-Synthetic constructs a shared data structure. By selecting specific data from the same frame, this structure supports the training requirements of various sub-tasks including depth completion, segmentation, upsampling, place recognition, and 3D reconstruction (See Tab. \ref{tab:briefintro}). 

Apart from supporting multiple tasks, we also introduce several essential settings that address key gaps in existing datasets. These settings enable the exploration of advanced hypotheses that are often limited by practical constraints, such as data collection costs. They provide valuable data for evaluating algorithm robustness and adaptability, such as domain adaptation and scene understanding, while also paving the way for further research. These features have also been validated through the following multi-task experiments, specifically in upsampling, segmentation, place recognition, and 3D reconstruction. In the cross-exploration of these tasks, these features have consistently ensured an improvement in task accuracy. The following sections will explain the implementation and significance of these features.

\noindent\textbf{Different density data in the same frame. }
In real-world scenarios, it is challenging to obtain data with varying point densities within a single frame due to the limited availability of LiDAR in vehicles. In our virtual environment, we strategically placed LiDAR with different channel counts (16/32/64/128 channels) at the same location. This unique configuration enables us to deliver, for the first time, scene-level point clouds with multiple densities in an upsampling task at the sensor level. The exploration of scene-level upsampling methods is crucial for the future development of intelligent point cloud processing. Additionally, we provide point clouds with varying densities for depth completion and semantic segmentation tasks. These can be used to assess the algorithm's robustness to density changes and for domain adaptation studies.

\noindent\textbf{Surface sampling from the city-level model. }
The existing scene-level data is obtained through LiDAR measurements, inevitably leading to the occurrence of obstructions during the collection process, making it challenging to capture the complete scene. To overcome this issue, we employ Monte Carlo Sampling \cite{hastings1970monte} to obtain point clouds and labels from the city mesh model's surface in our dataset. Subsequently, we divide the entire point cloud into blocks of 120m $\times$ 120m (consistent with the LiDAR detection distance), with a cropping interval of 6m. This approach ensures that the blocks simulate an unobstructed effect. By adopting this method, we aim to facilitate the exploration of techniques for scene-level completion of point clouds, while previous research primarily concentrates on completing individual objects. We also believe the city-level model can introduce innovative evaluation methods for city-scale 3D reconstruction.

\noindent\textbf{Simulate migration changes in environments. }
Traditional datasets are typically obtained through a one-time acquisition process, which does not account for temporal environmental variations like the progression of construction projects. While capturing these variations is important for developing adaptive algorithms, acquiring real-world datasets that span extended periods is a significant challenge, often requiring years of data collection. To explore this issue in a more controlled setting, we use a virtual environment to simulate short-term changes by making modifications to certain structures. Specifically, we design four routes with repetitive paths and temporary closures. Vehicles traversed these routes before and after the simulated changes, allowing us to collect datasets that reflect minor environmental shifts over a few days. Our preliminary findings suggested that even small-scale temporal and spatial changes can impact place recognition tasks. However, this work is only a preliminary step, providing a basic reference rather than a comprehensive solution to understanding the effects of environmental dynamics on such algorithms.

\section{Baseline Experiments}
In this section, we tested 5 typical tasks provided by WHU-Synthetic using mainstream methods. For a fair comparison, we faithfully followed the experimental settings of each baseline in the original publication and trained the models on our dataset. Unless otherwise noted, all experiments were done on a device with 64G RAM, and 1x 4070ti. From the experimental result, we found the positive effects of increased point cloud density on the accuracy of semantic segmentation and depth completion tasks, as well as the limitations of current upsampling algorithms for scene-level tasks, and the effect of environmental migration on the robustness of the network in the place recognition. These results validated the adaptability of the proposed dataset in various sub-tasks, and empirically demonstrated some conjectures. 

\begin{table}[!t]
    \centering
    \caption{Benchmarking results of depth completion on the test set. The accuracy is presented as root mean squared error (RMSE [mm]) and mean absolute error (MAE [mm])  $^{\dag}$We conduct experiments using data of 32/64/128-channel LiDAR point cloud projected image.}
    \label{tab:depth}
    \renewcommand\arraystretch{1.3}
    \resizebox{0.87\linewidth}{!}{
    \begin{tabular}{lccccc}
    \toprule
    \textbf{Method} & \textbf{Data}$^{\dag}$ & \textbf{RMSE $\downarrow$} & \textbf{MAE $\downarrow$} \\
    \midrule
    \multirow{4}{*}{NLSPN \cite{park2020non}} & 32 & 6972.50 & 4482.53 \\
    & 64 & 6538.66 & 4434.13 \\
    & 128 & 6021.36 & 3714.02 \\
    & KITTI & 741.68 & 199.59 \\
    \hline
    \multirow{4}{*}{PENet \cite{hu2021penet}} & 32 & \textbf{2833.97} & \textbf{803.90} \\
    & 64 & \textbf{2808.22} & \textbf{782.92} \\
    & 128 & \textbf{2724.75} & \textbf{739.43} \\
    & KITTI & 730.08 & 210.55 \\
    \hline
    \multirow{4}{*}{Physical Surface Mod \cite{zhao2021surface}} & 32 & 8683.49 & 5126.70\\
    & 64 & 8405.77 & 5001.68 \\
    & 128 & 8283.88 & 4950.97 \\
    & KITTI & 1239.84 & 298.30 \\
    \bottomrule
    \end{tabular}
    }
    \end{table}

\begin{figure}[!t]
	\centering
	\subfloat[32-channel completion]{\includegraphics[width=0.47\linewidth]{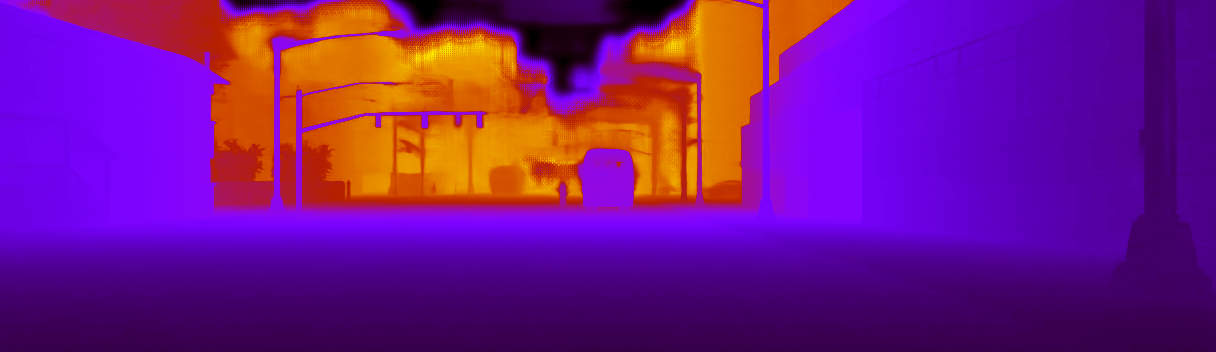}}
    \quad
	\subfloat[64-channel completion]{\includegraphics[width=0.47\linewidth]{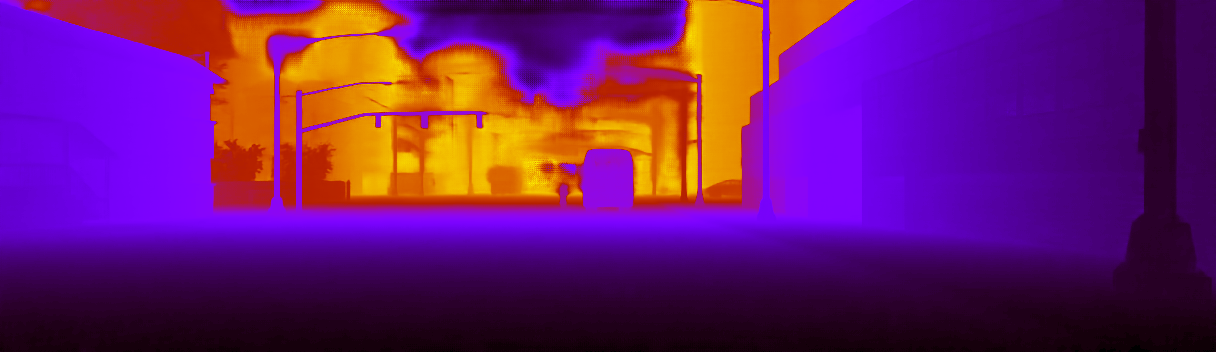}}
    \\
	\subfloat[128-channel completion]{\includegraphics[width=0.47\linewidth]{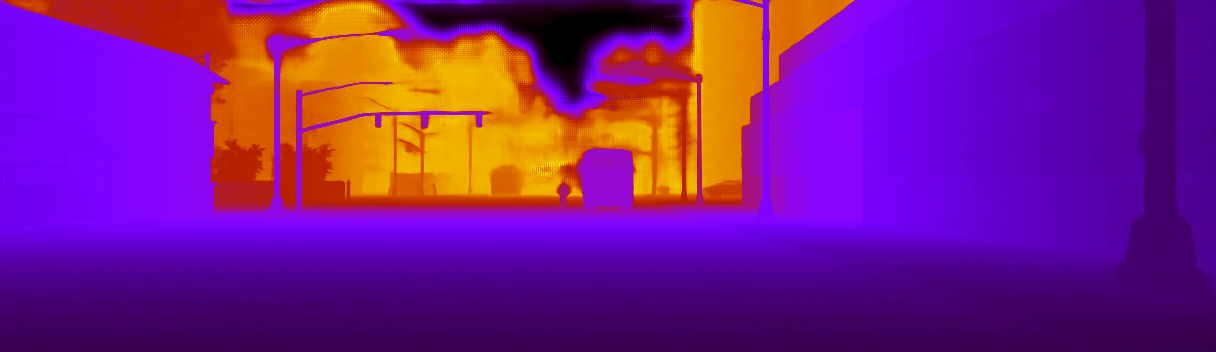}}
    \quad
	\subfloat[GT of WHU-Synthetic]{\includegraphics[width=0.47\linewidth]{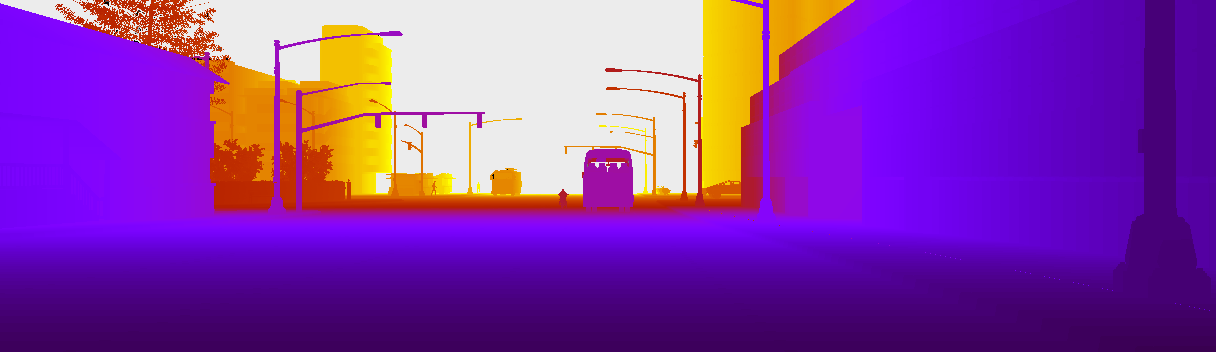}}
	\\
	\subfloat[RGB]{\includegraphics[width=0.47\linewidth]{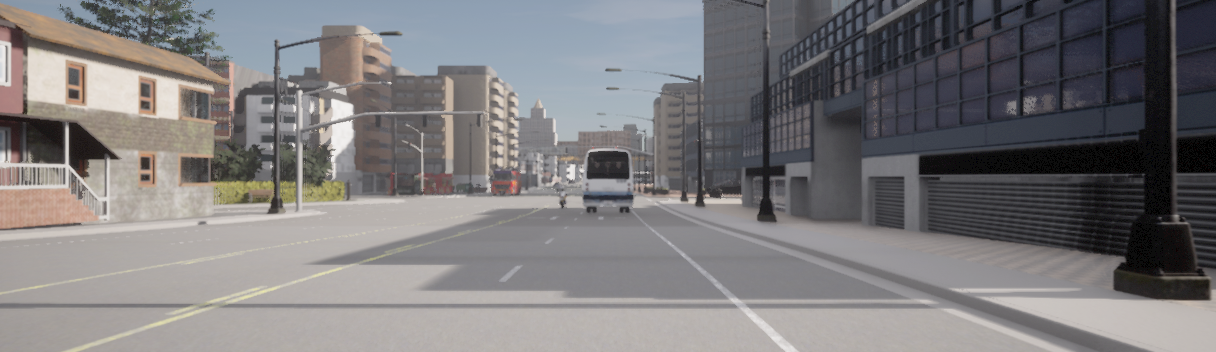}}
    \quad
	\subfloat[GT of KITTI DC]{\includegraphics[width=0.47\linewidth]{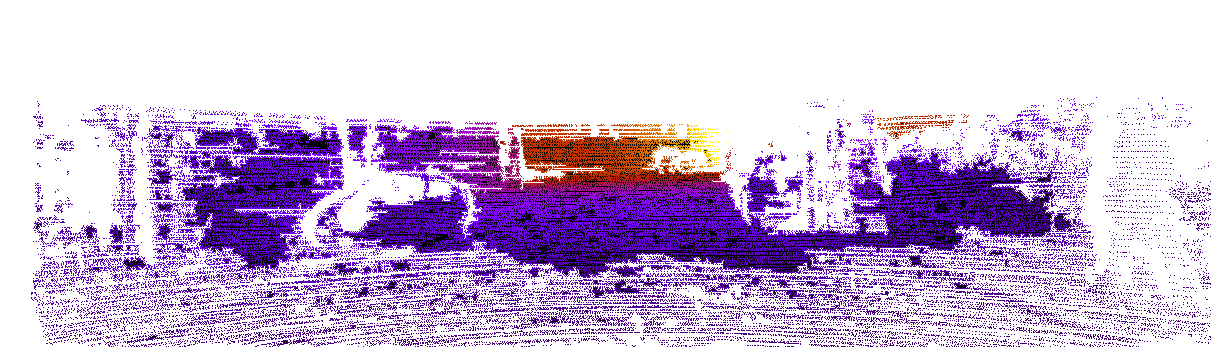}}
	\\
	\caption{Depth completion results of PENet from different input (32/64/128-channel LiDAR projected image).}
 \label{fig:depth}
\end{figure}

\noindent\textbf{Depth completion. }
We carried out experiments on the projected image of different channel counts using learning-based methods \cite{park2020non, hu2021penet} and traditional methods \cite{zhao2021surface}. PENet testing was done on 1x RTX 3090. Tab. \ref{tab:depth} presents our experimental results. PENet achieved the best results, while the other two methods performed poorly. However, it should be noted that our ground truth is generated by the renderer and is denser than the annotated depth maps in the KITTI DC dataset, so the accuracy of these methods on our dataset is much lower. This presented a more challenging baseline, thereby driving the development of algorithms. Additionally, we observed that as the LiDAR channel count increased, the accuracy of all methods improved to varying degrees, which is particularly evident in NLSPN. This displayed the limitations of existing methods and demonstrates the enhancing effect of increasing point density on depth completion tasks. Visualization results (Fig. \ref{fig:depth}) showed the outcome of PENet. It could be observed from both quantitative and visual results that with the increase in point density, there is a slight improvement in the accuracy of the completion results. Meanwhile, this multi-density setup can effectively support research in domain adaptation and transfer.

\noindent\textbf{LiDAR odometry estimation. }
We conducted experiments on three sequences according to the settings of KITTI Odometry. Tab. \ref{tab:odom} presented our experimental results, where KISS-ICP \cite{vizzo2023ral} achieved the best results in Town05 and Town10, while A-LOAM \cite{zhang2014loam} achieved the best results in Town03. However, it was unusual that MULLS \cite{pan2021mulls} exhibits significant deviations in Town03 and Town05, and A-LOAM's results in Town05 were also unsatisfactory. We also tested ORB-SLAM2 \cite{murORB2} using stereo images, but the final performance was poor, as the algorithm lost track early in the sequence. Overall, KISS-ICP exhibited the strongest robustness, which aligns with its characteristic of not requiring parameter tuning. In addition, the city-level models provided by WHU-Synthetic can offer a new testing benchmark for the entire 3D reconstruction task, thereby driving the development of more adaptive methods.

\begin{table}[ht]
\centering
\caption{Benchmarking results of LiDAR odometry estimation on each sequence. The errors are presented as absolute trajectory error (ATE [m]) and absolute rotational error (ARE [rad]).}
\label{tab:odom}
\renewcommand\arraystretch{1.3}
\resizebox{0.73\linewidth}{!}{
\begin{tabular}{lccccc}
\toprule
\textbf{Method} & \textbf{Sequence} & \textbf{ATE $\downarrow$} & \textbf{ARE $\downarrow$}  \\
\midrule
\multirow{3}{*}{KISS-ICP \cite{vizzo2023ral}} & Town03 & 6.399 & 0.052 \\
 & Town05 & \textbf{9.583} & \textbf{0.069} \\
 & Town10 & \textbf{3.124} & \textbf{0.022} \\
\hline
\multirow{3}{*}{A-LOAM \cite{zhang2014loam}} & Town03 & \textbf{4.514} & \textbf{0.038} \\
 & Town05 & 69.811 & 0.402\\
 & Town10 & 3.733 & 0.034 \\
 \hline
\multirow{3}{*}{MULLS \cite{pan2021mulls}} & Town03 &   26.445& 0.172\\
 & Town05 & 228.50 & 1.221\\
 & Town10 & 5.224 & 0.064 \\
\bottomrule
\end{tabular}
}
\end{table}

\noindent\textbf{Point cloud place recognition. }
Our place recognition dataset used the same data preprocessing procedure as benchmark datasets proposed in \cite{uy2018pointnetvlad} (except for removing ground plane points). To investigate the impact of dynamic environments, we generated four sequences before and after simulating environmental changes, denoted as Org. and Mig. respectively. Each run was split into two disjoint reference maps used for training and testing, and each reference map was organized as a set of sub maps at regular intervals. We choose PointNetVLAD \cite{uy2018pointnetvlad}, PPTNet \cite{hui2021pptnet}, and MinkLoc3dv2 \cite{9956458} as global feature extractors for place recognition and train each extractor with 2 different training sets of sub maps from Org. only and Org.+Mig. Regarding validation, we uniformly employed cross-validation on all eight sequences. Tab. \ref{tab:recognition} demonstrated that training PPTNet \cite{hui2021pptnet} and MinkLoc3dv2 \cite{9956458} with two sets of sequence data lead to noticeable improvement compared to using only the original sequence. This finding highlighted the influence of environmental changes on the accuracy of place recognition for the first time.

\begin{table}[ht]
\centering
\caption{Benchmarking results of place recognition. Average recall at 1(AR@1 [\%]) and at 5(AR@5 [\%]) are reported. $^{\dag}$Org. (Original) denotes training that exclusively utilizes data from the original environment, while Org. (Original)+Mig. (Migrated) indicates training that incorporates data from both the original and migrated environments.}
\label{tab:recognition}
\renewcommand\arraystretch{1.3}
\resizebox{0.84\linewidth}{!}{
\begin{tabular}{lccc}
\toprule
\textbf{Method} & \textbf{Data}$^{\dag}$ & \textbf{AR@1 $\uparrow$} & \textbf{AR@5 $\uparrow$} \\
\midrule
\multirow{2}{*}{PointNetVLAD \cite{uy2018pointnetvlad}} & Org. & 87.7 & 90.4 \\
 & Org.+Mig. & 84.3 & 90.3 \\
\hline
\multirow{2}{*}{PPTNet \cite{hui2021pptnet}} & Org. & \textbf{92.6} & 94.4 \\
 & Org.+Mig. & 94.1 & 95.0 \\
\hline
\multirow{2}{*}{MinkLoc3dv2 \cite{9956458}} & Org. & 89.9 & \textbf{94.9} \\
 & Org.+Mig. & \textbf{97.1} & \textbf{98.5}  \\
\bottomrule
\end{tabular}
}
\end{table}

\begin{figure*}[ht]
	\centering
	\subfloat[16-$>$128]{\includegraphics[width=0.23\linewidth]{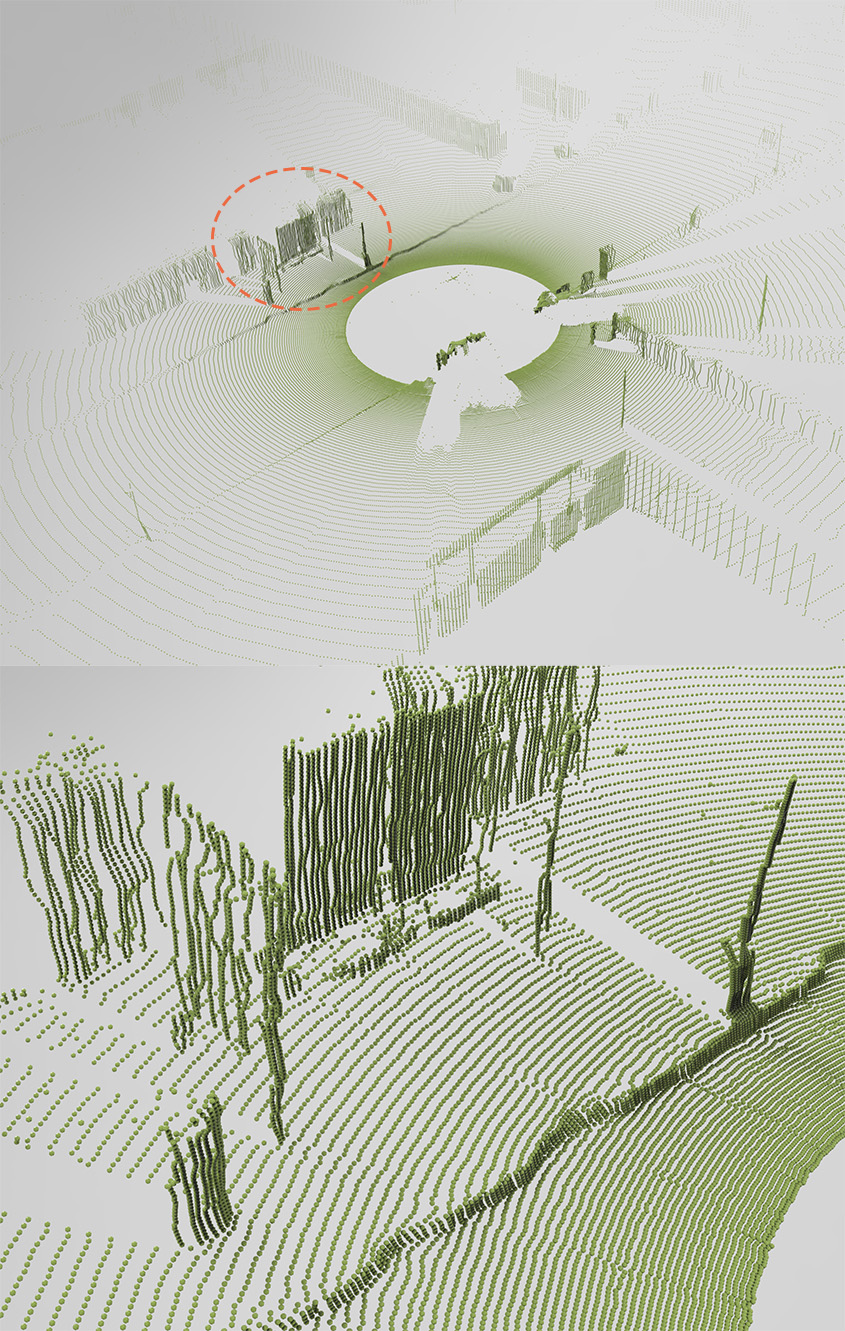}}
    \hfil
	\subfloat[32-$>$128]{\includegraphics[width=0.23\linewidth]{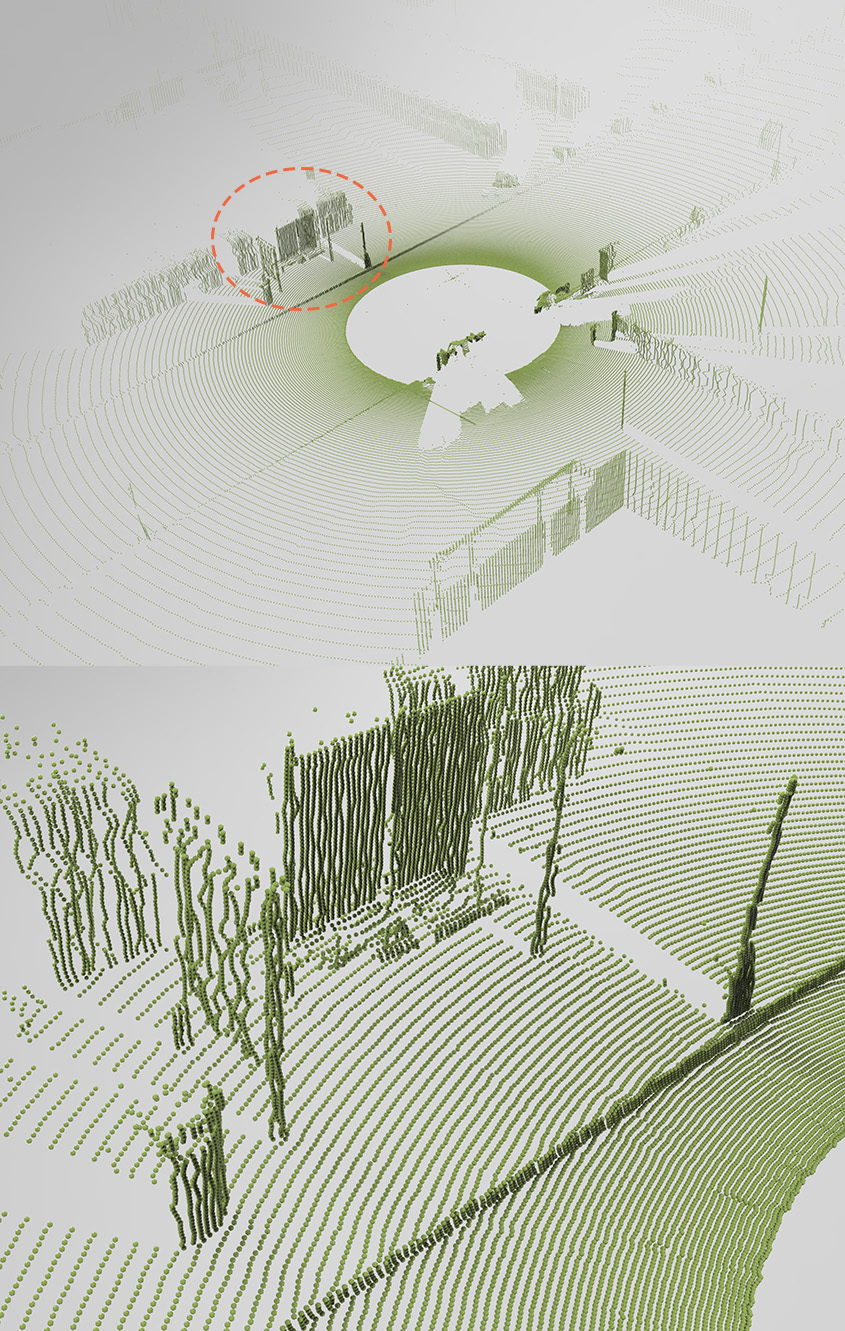}}
    \hfil
	\subfloat[64-$>$128]{\includegraphics[width=0.23\linewidth]{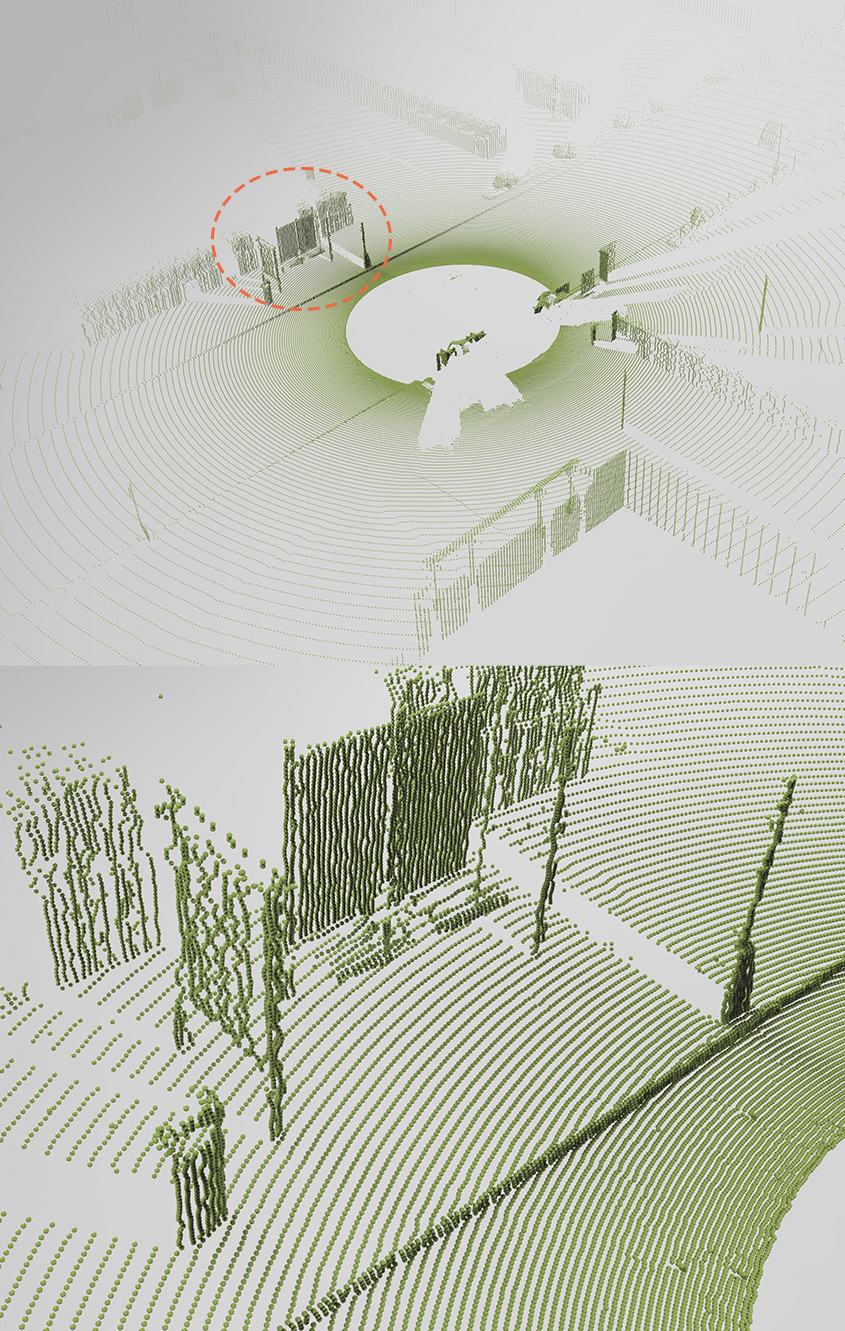}}
    \hfil
	\subfloat[GT]{\includegraphics[width=0.23\linewidth]{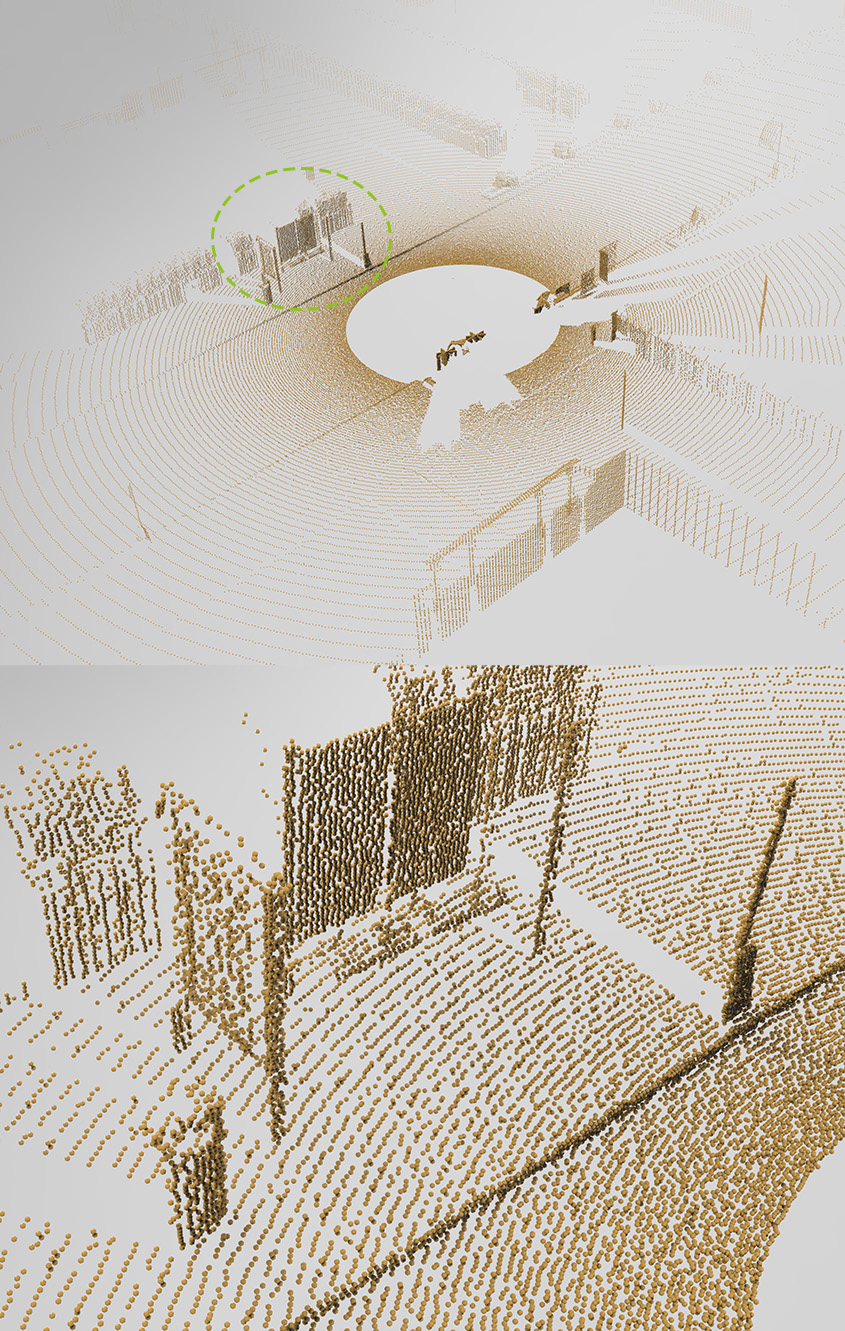}}
	\caption{Upsampling results of ILN from different inputs (16/32/64-channel LiDAR scan). The ground truth is the 128-channel LiDAR scan.}
 \label{fig:ups}
\end{figure*}

\noindent\textbf{Scene-level point cloud upsampling. }
We conducted scene-level upsampling experiments using interpolation-based \cite{kwon2022implicit} and transformer-based \cite{yang2024tulip} methods. These experiments were done on 1x RTX3090. The results are presented in Tab. \ref{tab:upsample}. It is important to mention that HD metrics do not effectively reveal the superiority or inferiority of the methods in scene-level upsampling since HD only reflects the maximum mismatch between point sets. As can be seen, ILN exhibited higher accuracy than TULIP-L on each task, which is contrary to the results provided in the original paper. We speculated that this may be due to the noise added to our dataset. The specific visual results of ILN are shown in Fig. \ref{fig:ups}. With the continuous increase in input density, the accuracy of both methods improved. From the visualization results, upsampling can effectively restore the details of the point cloud. The multi-density setup of our dataset offers new perspectives for point cloud data augmentation methods under real-world conditions.

\begin{table}[ht]
\centering
\caption{Benchmarking results of scene-level point cloud upsampling on the test set. Chamfer distance (CD) and Hausdorff distance (HD) are reported as evaluation metrics. Note that CD and HD are multiplied by 10$^3$. $^{\dag}$We use 16/32/64-channel LiDAR scan as input and 128-channel LiDAR scan as ground truth. }
\label{tab:upsample}
\renewcommand\arraystretch{1.3}
\resizebox{0.73\linewidth}{!}{
\begin{tabular}{lccccc}
\toprule
\textbf{Method} & \textbf{Data$^{\dag}$}& \textbf{CD $\downarrow$} & \textbf{HD $\downarrow$}\\
\midrule
\multirow{3}{*}{ILN \cite{kwon2022implicit}} & 16-$>$128 & \textbf{0.052} & \textbf{101.435} \\
 & 32-$>$128 & \textbf{0.016} & \textbf{76.715} \\
 & 64-$>$128 & \textbf{0.007} & \textbf{62.349} \\
\hline
\multirow{3}{*}{TULIP-L \cite{yang2024tulip}} & 16-$>$128 & 0.122 & 177.328 \\
 & 32-$>$128 & 0.048 & 135.163 \\
 & 64-$>$128 & 0.014 & 85.945 \\
\bottomrule
\end{tabular}
}
\end{table}

\begin{table*}[t]
\centering
\caption{Benckmarking results of point cloud semantic segmentation on the validation set. We conducted experiments on data with different point densities. Mean IoU (mIoU [\%]), and per-class IoU [\%] scores are reported.}
\label{tab:seg}
\renewcommand\arraystretch{1.3}
\resizebox{\linewidth}{!}{
\begin{tabular}{lcccccccccccccccccc}
\toprule
\textbf{Method} & \textbf{Data} & \textbf{mIoU $\uparrow$} & \rotatebox{90}{car} & \rotatebox{90}{bicycle} & \rotatebox{90}{motorcycle} & \rotatebox{90}{truck} & \rotatebox{90}{other-vehicle} & \rotatebox{90}{person} & \rotatebox{90}{rider} & \rotatebox{90}{road} & \rotatebox{90}{sidewalk} & \rotatebox{90}{ground} & \rotatebox{90}{building} & \rotatebox{90}{fence} & \rotatebox{90}{vegetation} & \rotatebox{90}{terrain} & \rotatebox{90}{pole} & \rotatebox{90}{traffic-sign} \\
\midrule
\multirow{3}{*}{Minkowski U-Net \cite{choy20194d}} & 32 & \textbf{69.0} & 81.8 & 41.7 & 51.1 & 86.1 & 73.1 & 60.6 & 70.3 & 94.0 & 75.8 & 53.4 & 78.9 & 80.0 & 79.5 & 55.6 & 57.8 & 63.9 \\
 & 64 & \textbf{77.2} & 90.1 & 56.2 & 58.4 & 92.7 & 81.6 & 76.1 & 79.8 & 93.0 & 83.4 & 64.9 & 89.4 & 88.0 & 90.7 & 57.7 & 63.5 & 68.9 \\
 & 128 & \textbf{82.9} & 95.0 & 73.9 & 76.4 & 95.0 & 76.3 & 84.3 & 82.3 & 93.5 & 86.6 & 63.9 & 95.8 & 93.6 & 95.9 & 60.0 & 80.3 & 73.4 \\
 \hline
\multirow{3}{*}{RandLA-Net \cite{hu2019randla}} & 32 & 60.2
& 49.1 & 24.0 & 35.3 & 91.3 & 33.8 & 26.6 & 18.4 & 97.0 & 88.8 & 63.5 & 94.3 & 88.9 & 92.5 & 78.6 & 81.0 & 0.0 \\
 & 64 & 70.7 & 66.9 & 26.0 & 53.0 & 96.7 & 82.2 & 56.6 & 21.8 & 98.4 & 92.0 & 73.0 & 97.5  & 94.2 & 96.6 & 87.7 & 89.0 & 0.0 \\
 & 128 & 73.2 & 71.0 & 34.1 & 59.2 & 96.9 & 81.7 & 68.8 & 31.0 & 98.5 & 92.3 & 68.2 & 97.6 & 94.9 & 97.1 & 88.3 & 91.5 & 0.0 \\
 \hline
\multirow{3}{*}{KPConv \cite{thomas2019KPConv}} & 32 & 62.3 & 90.7 & 42.0 & 48.9 &82.8& 1.4 &73.5 &65.9 &90.8 &73.3 &8.6 &80.0 &63.7 &86.9 &43.5 &71.2 &72.9\\
 & 64 & 76.2 &95.8 &64.1 &65.5 &91.4 &27.7 &84.4 &76.8 &92.8 &80.4 &49.7 &91.8 &87.9 &94.4 &53.3 &80.5 &82.7  \\
 & 128 & 77.7 & 92.7 & 68.7 & 67.6&96.0&90.7&87.8&79.7&90.8&74.7&32.7&88.2&83.3&96.3&41.3&73.8&79.3 \\
\bottomrule
\end{tabular}
}
\end{table*}

\begin{figure*}[ht]
	\centering
	\subfloat[32-channel pred]{\includegraphics[width=0.31\linewidth]{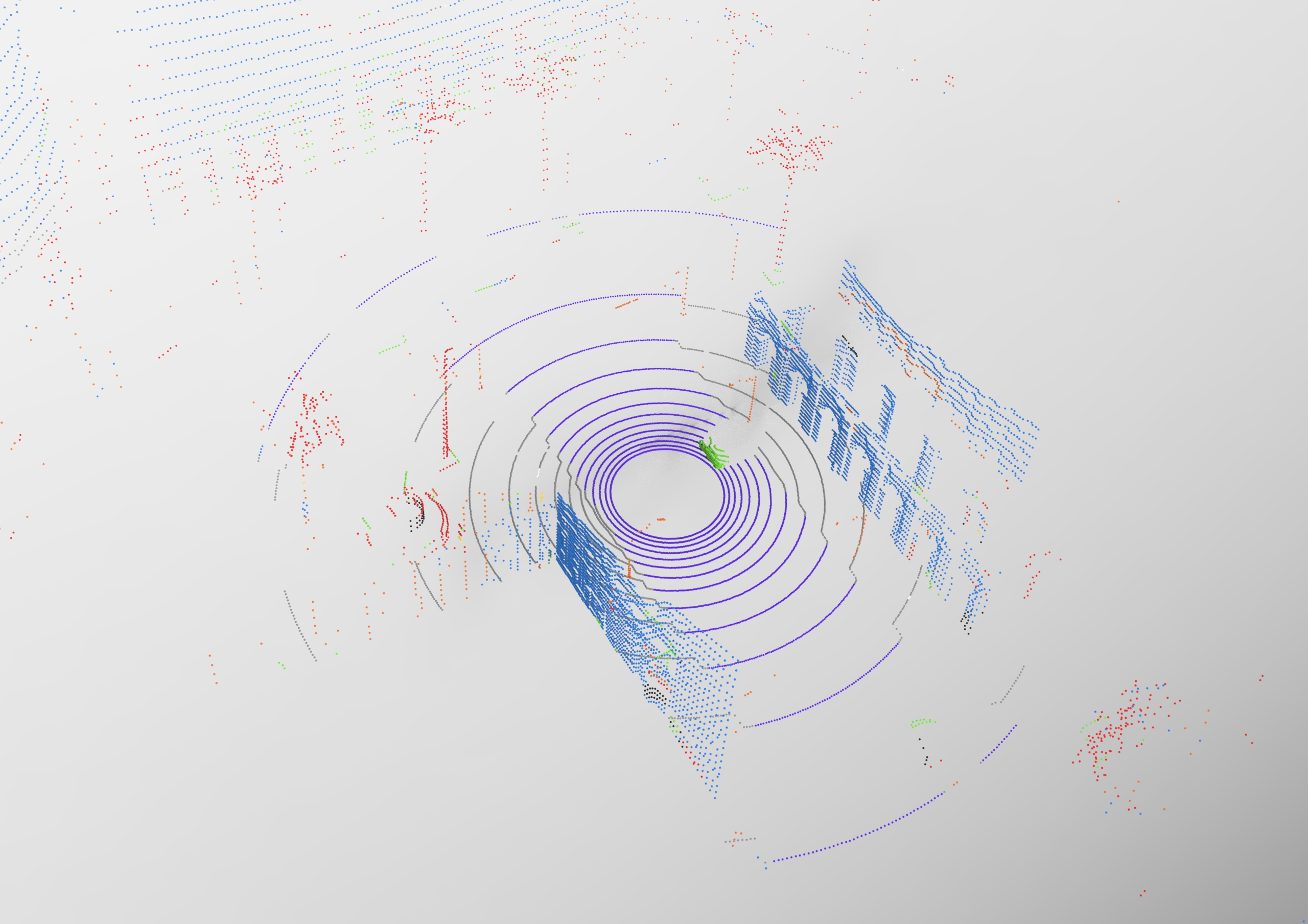}}
    \hfil
	\subfloat[64-channel pred]{\includegraphics[width=0.31\linewidth]{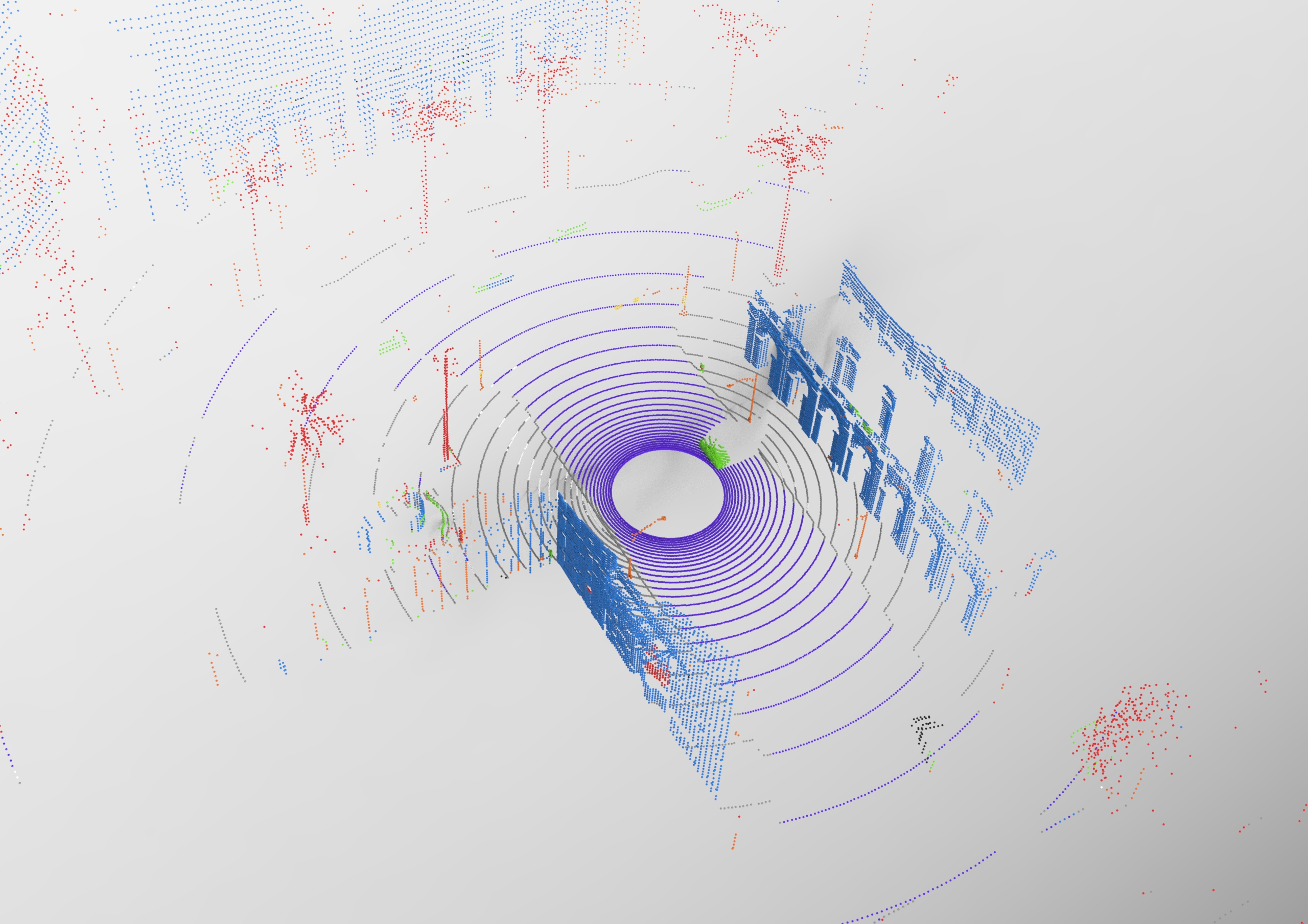}}
    \hfil
	\subfloat[128-channel pred]{\includegraphics[width=0.31\linewidth]{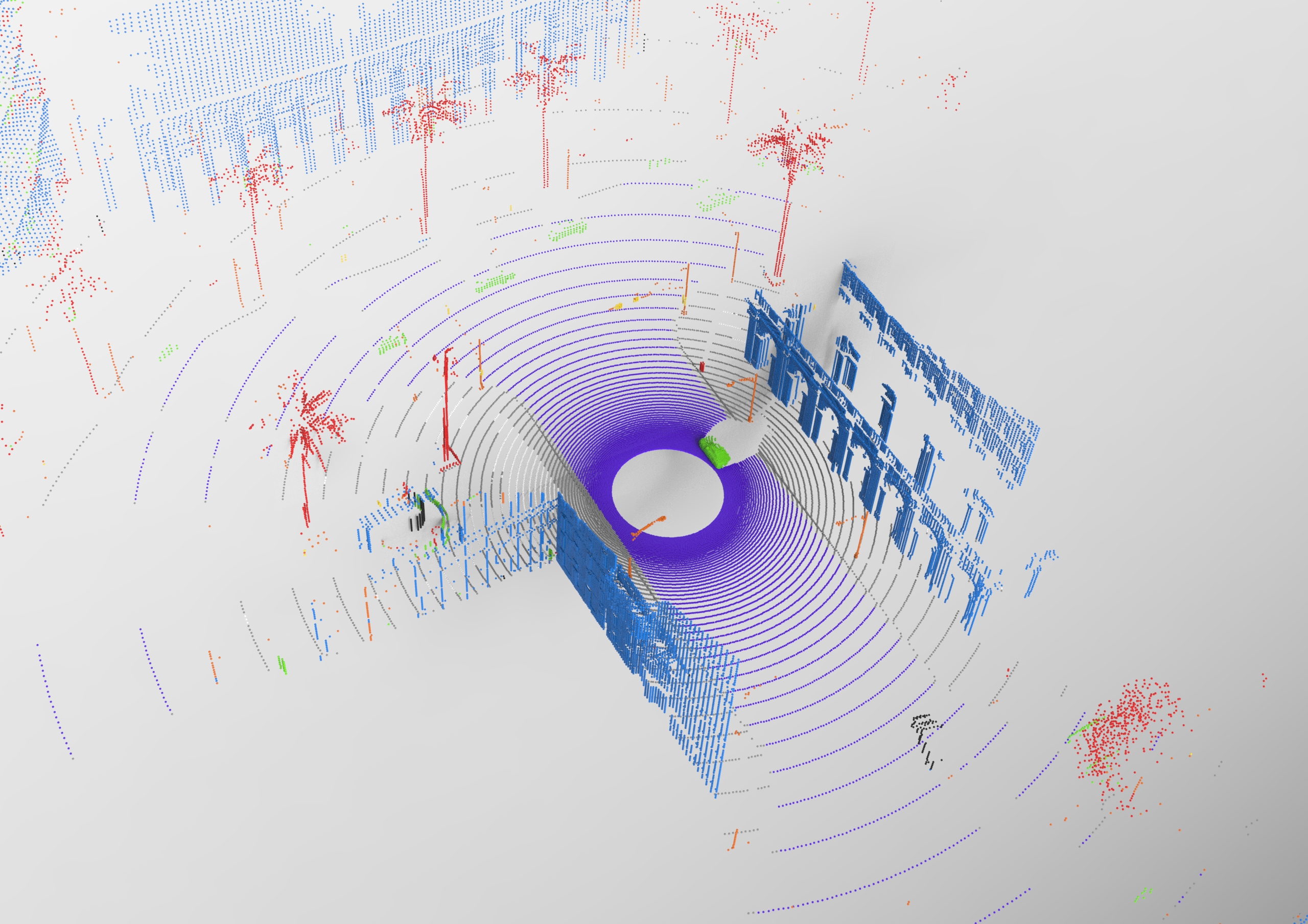}}
    \\
	\subfloat[32-channel GT]{\includegraphics[width=0.31\linewidth]{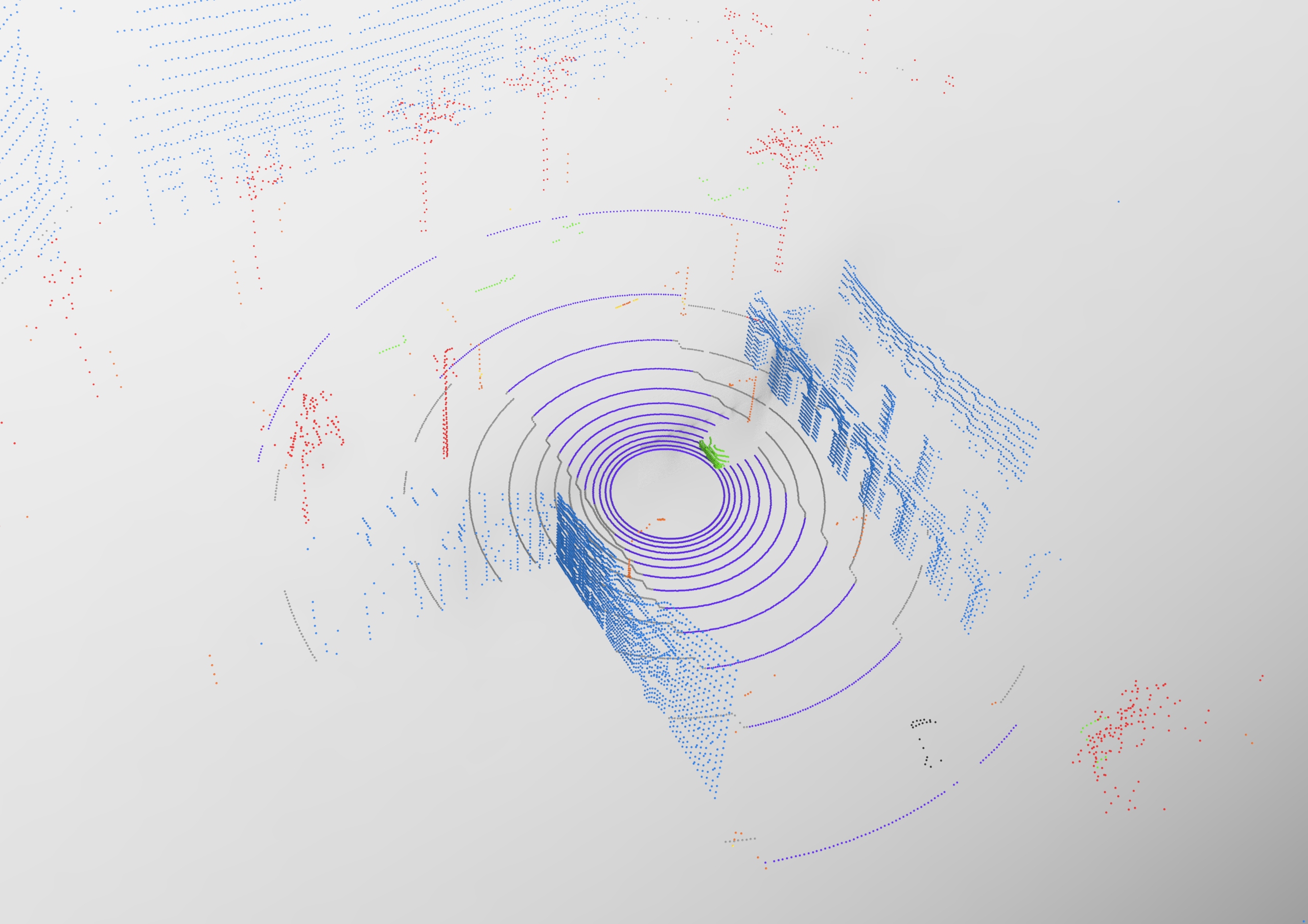}}
    \hfil
	\subfloat[64-channel GT]{\includegraphics[width=0.31\linewidth]{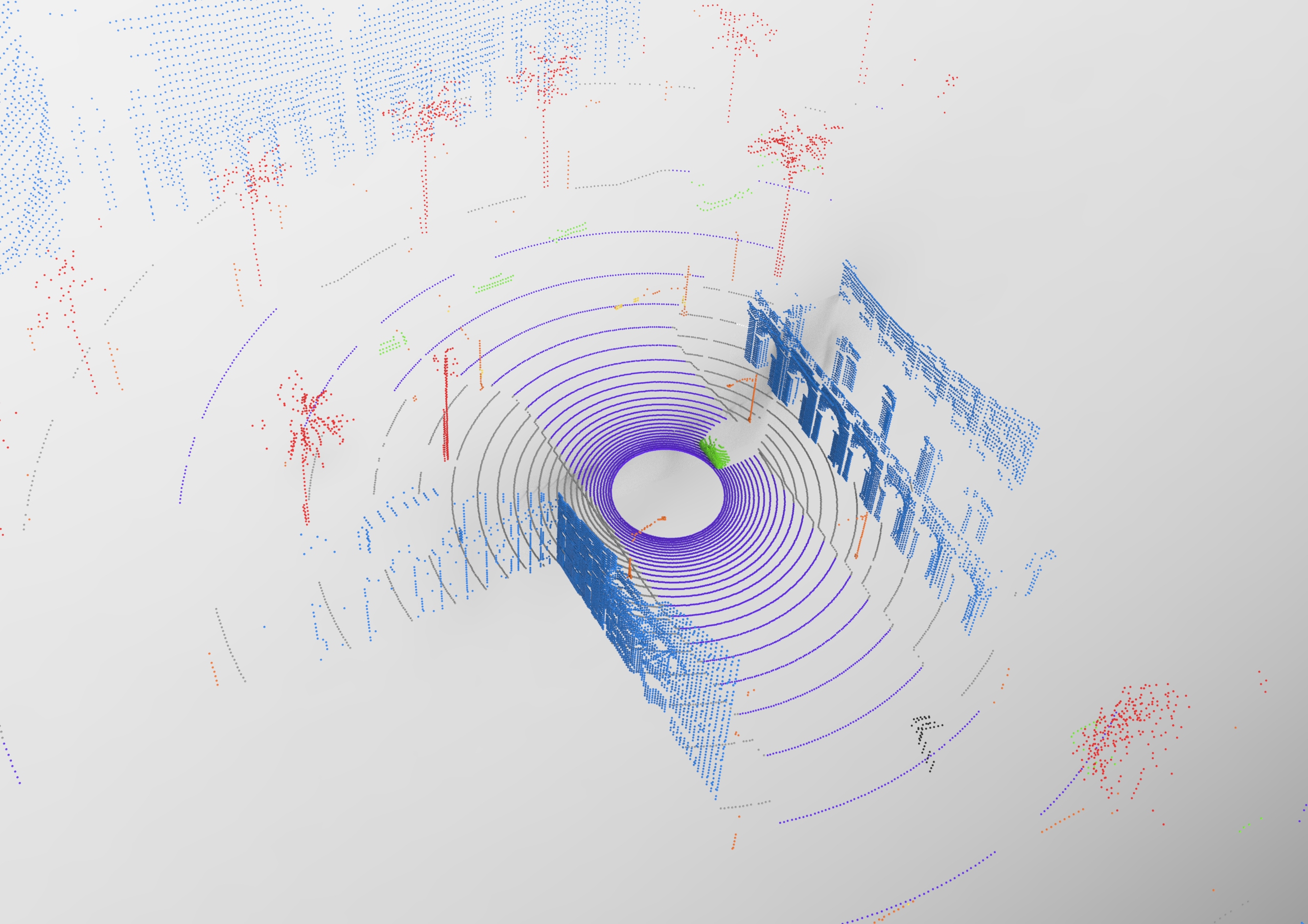}}
    \hfil
	\subfloat[128-channel GT]{\includegraphics[width=0.31\linewidth]{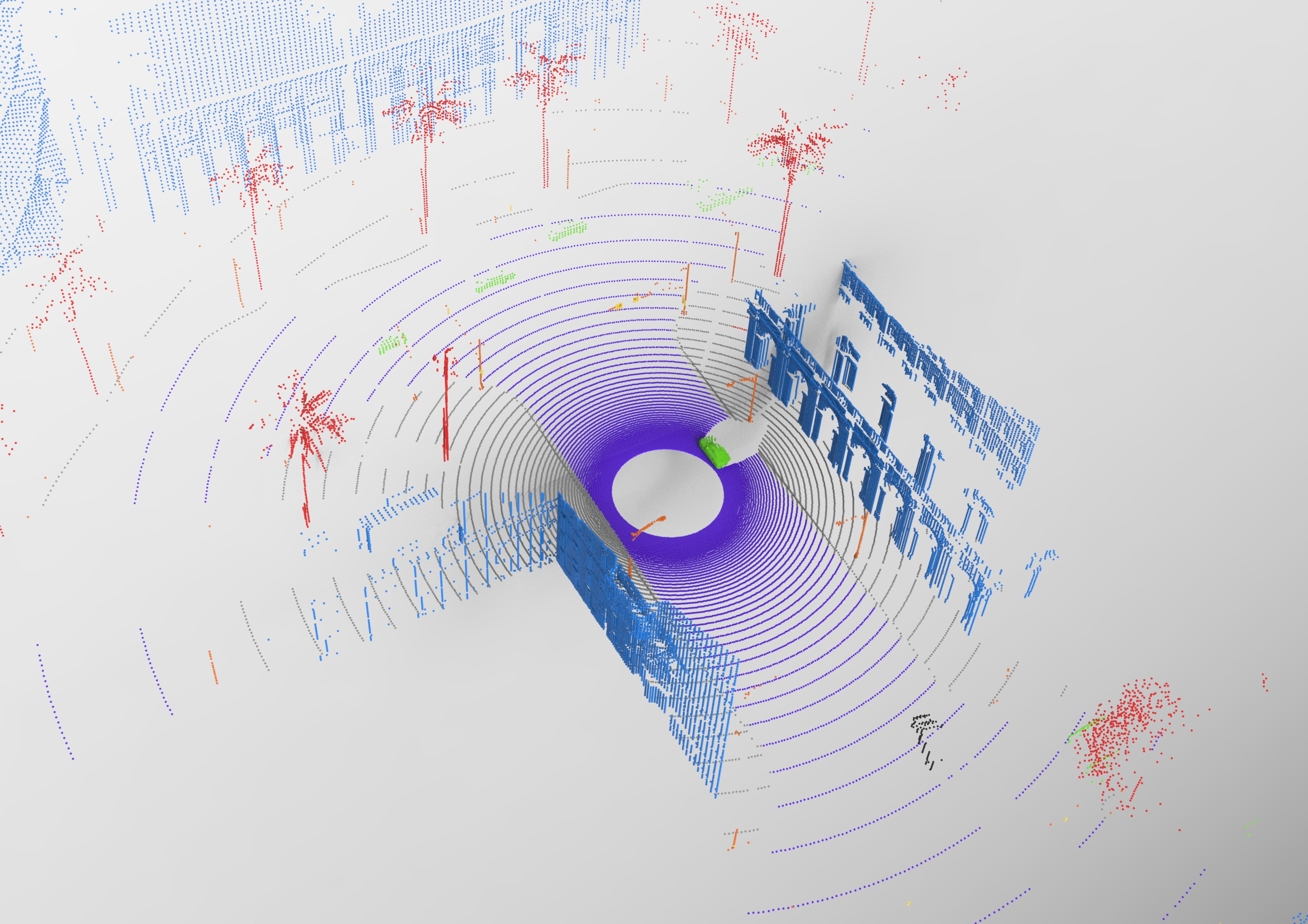}}
	\caption{Semantic segmentation results of Minkowski U-Net from different inputs (32/64/128-channel LiDAR scan). }
 \label{fig:seg}
\end{figure*}

\noindent\textbf{Point cloud semantic segmentation. }
We utilized pointwise-MLP-based\cite{hu2019randla}, point-convolution-based\cite{thomas2019KPConv} and voxel-based\cite{choy20194d} methods to test our dataset. The results were shown in Tab. \ref{tab:seg} and Fig. \ref{fig:seg}. The point-convolution-based Minkowski U-Net achieved the best results (voxel size is all set to 0.05). As for the pointwise-MLP-based RandLA-Net, the test results for different channel counts were not satisfactory. For example, for small objects like traffic signs, there were even cases where the IoU was 0. We speculate that this is because the downsampling strategy used in RandLA-Net is not suitable for segmenting small objects in large scenes, resulting in performance degradation. In point-convolution-based KPConv, compared to the 64-channel data, the 128-channel data did not show a significant improvement. This was because, for the KPConv, with higher-density input point clouds, precise adjustment of the neighborhood radius is essential to prevent performance degradation \cite{thomas2019KPConv}. The multi-density point clouds we offer facilitate robustness testing of segmentation methods across domains, aiding researchers in exploring domain adaptation or generalization studies in real-world conditions.

\section{Domain Shift Experiments}
To quantitatively explore the domain shift between our synthetic dataset and real-world datasets, we tested the pre-trained models on mainstream real-world data across learning-based tasks: depth completion, upsampling, and semantic segmentation.

\noindent\textbf{Depth completion.} We trained the E-Net model on the 64-channel LiDAR projection images from WHU-Synthetic for 50 epochs. After training, we tested the model on the KITTI DC dataset to observe the impact of domain shift. As shown in Tab. \ref{tab:domain_com}, the model achieved an RMSE of 2808.22 and an MAE of 782.92 on the WHU-Synthetic dataset. When the same model was tested on the KITTI DC test set, the RMSE increased to 3608.65 and the MAE to 996.50. The results suggested that the WHU-Synthetic dataset has the potential to assist in the training of depth completion models, as it allowed the model to generalize reasonably well on KITTI DC.

\begin{table}[ht]
\centering
\caption{Domain adaptation results of depth completion on E-Net.}
\label{tab:domain_com}
\renewcommand\arraystretch{1.3}
\resizebox{0.6\linewidth}{!}{
\begin{tabular}{llll}
\toprule
\textbf{Dataset} & \textbf{RMSE $\downarrow$} & \textbf{MAE $\downarrow$} \\
\midrule
WHU-Synthetic & 2808.22 & 782.92 \\
KITTI DC & 3608.65 & 996.50 \\
\bottomrule
\end{tabular}
}
\end{table}

\begin{figure}[ht]
	\centering 
	\subfloat[KITTI input]{\includegraphics[width=0.48\linewidth]{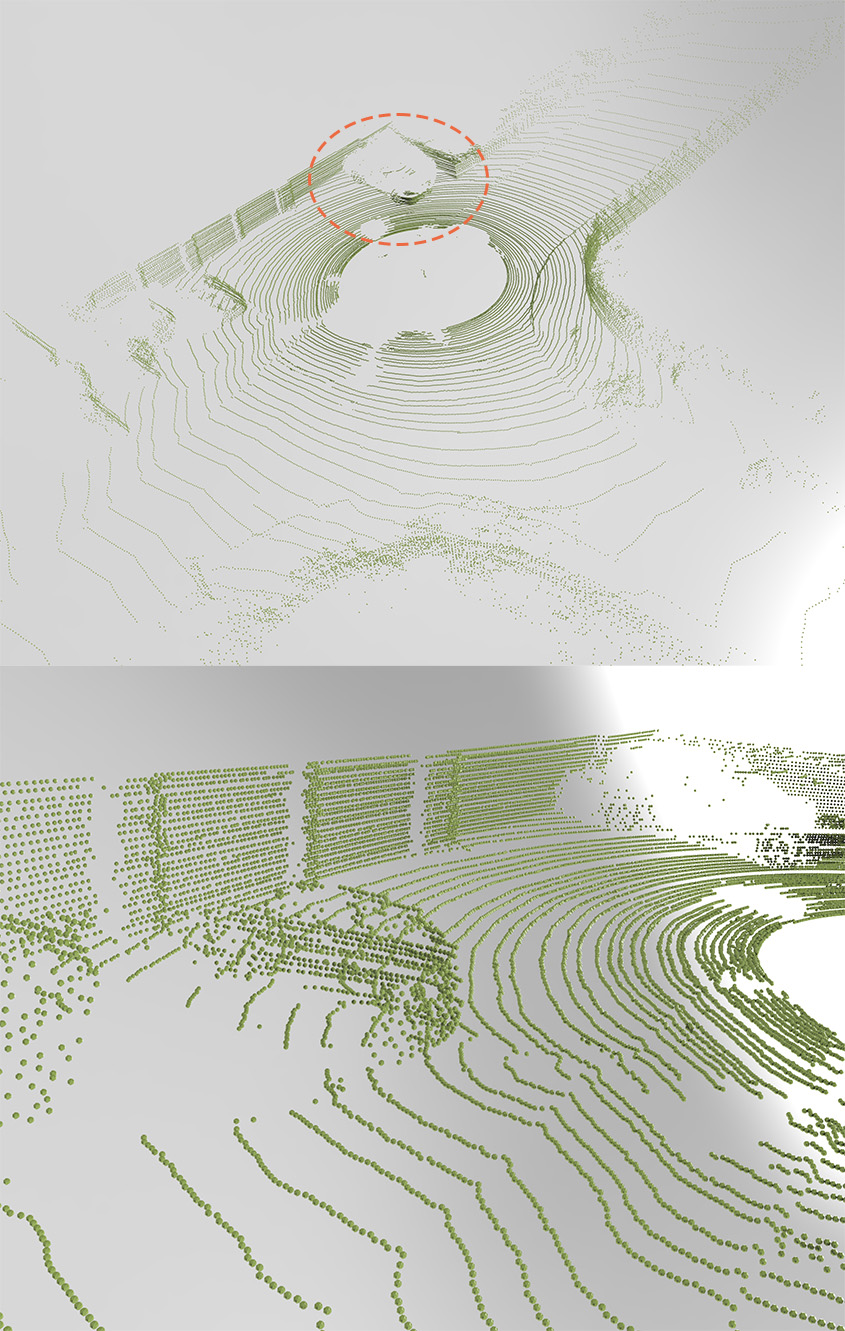}}
    \hfil
	\subfloat[ILN prediction]{\includegraphics[width=0.48\linewidth]{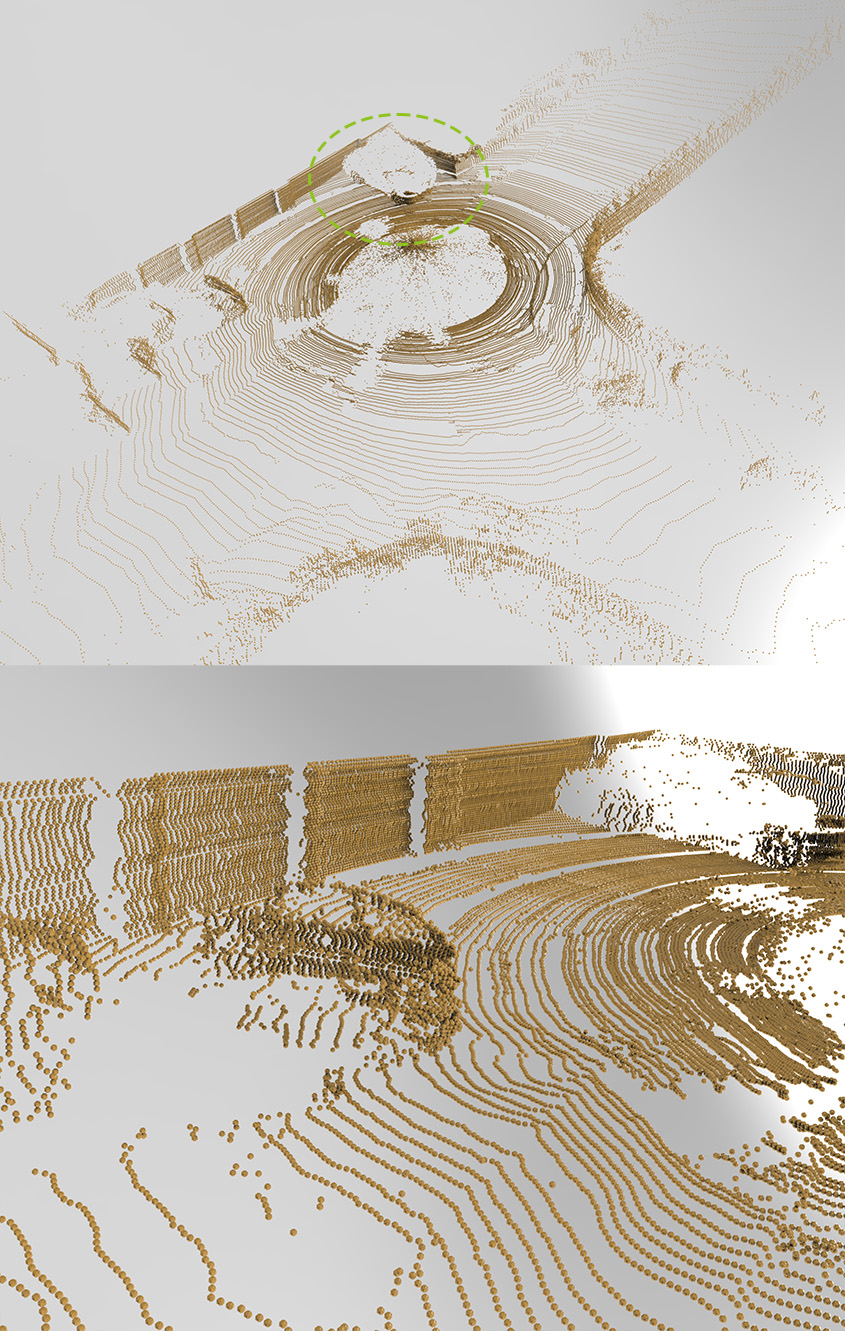}}
	\caption{Domain adaptation visualization results of upsampling on ILN.}
\label{fig:domain_ups}
\end{figure}

\noindent\textbf{Scene-level point cloud upsampling}. Since there is currently no real-world scene-level upsampling dataset, we could not directly compare the differences in accuracy. We trained ILN on the WHU-Synthetic dataset and tested it on the KITTI dataset, upsampling the 64-channel point cloud to 128-channel. Fig. \ref{fig:domain_ups} showed the visualization result of one frame. The network trained on WHU-Synthetic could effectively enhance the KITTI data. Although the final result introduces some noise, it also improved the details of the point cloud to some extent, reconstructed the local areas of small objects, and exhibited a relatively small domain shift.

\noindent\textbf{Point cloud semantic segmentation.} We trained the Minkowski U-Net model on the 64-channel LiDAR data from WHU-Synthetic for 50 epochs. We used random rotation along the z-axis was performed for the data augmentation. Subsequently, we tested the trained model on the Semantic KITTI dataset. For this experiment, we focused on the 9 classes that are common to both datasets, while ignoring the other classes. The results of this experiment were presented in Tab. \ref{tab:domain_seg}.

\begin{table}[ht]
\centering
\caption{Domain adaptation results of semantic segmentation on Minkowski U-Net.}
\label{tab:domain_seg}
\renewcommand\arraystretch{1.3}
\resizebox{\linewidth}{!}{
\begin{tabular}{lccccccccccccccccccc}
\toprule
 \textbf{Dataset} & \textbf{mIoU $\uparrow$} & \rotatebox{90}{car} & \rotatebox{90}{person} & \rotatebox{90}{road} & \rotatebox{90}{sidewalk} & \rotatebox{90}{building} & \rotatebox{90}{fence} &  \rotatebox{90}{pole} & \rotatebox{90}{vegetation} & \rotatebox{90}{traffic-sign} \\
\midrule 
WHU-Synthetic &83.1 & 90.1 & 81.6 &93.0 &83.4 &89.4 &88.0 &63.5 &90.7 &68.9 \\
SemanticKITTI &33.8 & 57.5 & 34.6 &46.8 &26.6 &37.6 &1.2 &25.1 &69.6 &5.5  \\
\bottomrule
\end{tabular}
}
\end{table}

The model trained on WHU-Synthetic achieved a mIoU of 33.8\% when tested on Semantic KITTI. The performance of specific classes varied, with some classes showing smaller domain shifts than others. For instance, the model achieved an IoU of 57.5\% for the 'car' class and 69.6\% for the 'vegetation' class, indicating relatively small domain shifts for these categories. In contrast, classes such as 'fence' and 'traffic-sign' exhibited larger domain shifts, with IoUs of 1.2\% and 5.5\%, respectively.

The relatively high IoU scores for 'car' and 'vegetation' suggested that the features learned from the WHU-Synthetic dataset generalize well to the Semantic KITTI dataset for these categories. This could be attributed to the similar appearance and structure of these objects across both datasets. On the other hand, the larger domain shifts observed in classes like 'fence' and 'traffic-sign' may be due to differences in the representation and frequency of these objects between the two datasets.

\section{Multi-task Exploration}
In this section, we designed a framework with a shared backbone based on the common feature extraction requirements of the upsampling and segmentation tasks. Meanwhile, we also tested some current methods that use auxiliary tasks to support the main task, to explore the mutually beneficial relationships between different subtasks. This included place recognition and reconstruction, as well as loop closure detection and segmentation. The experiments verified that there is indeed a mutually reinforcing effect between the two subtasks, and simultaneous training accelerates the convergence and improves the performance. This also partially demonstrated that our dataset can facilitate 3D multi-task research and represents more possibilities for future research.

\begin{figure*}[th]
    \centering
    \includegraphics[width=\linewidth]{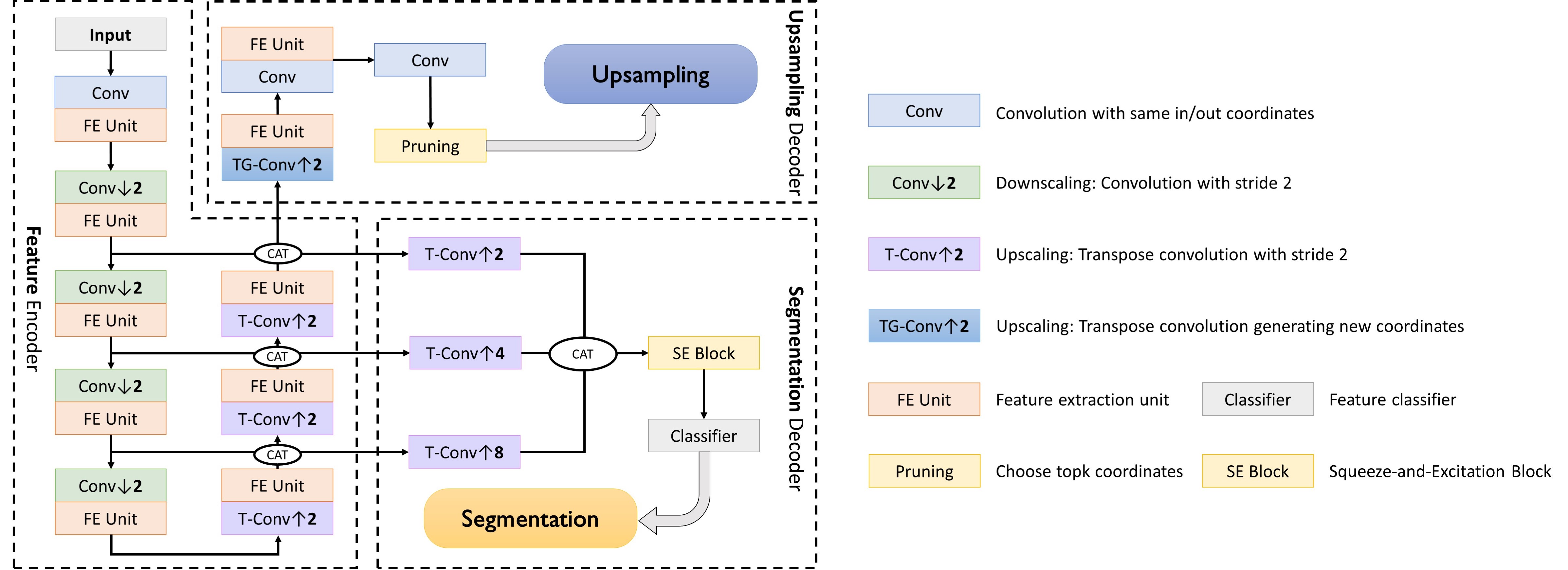}
    \caption{Multi-task network architecture. We adopt PU-Dense as the backbone network architecture and expand it with a semantic segmentation branch.}
    \label{fig:network}
\end{figure*}

\subsection{Upsampling \& Segmentation}
The labeled point cloud with different densities provided by WHU-Synthetic facilitates improved implicit surface learning. This enables data augmentation at the sensor level, while the generated digital assets (mesh models) can easily expand the range of segmentation categories, offering support for more fine-grained information.

\subsubsection{The Network}
We adopted PU-Dense \cite{akhtar2022pu} network architecture as the backbone network and extended a branch for semantic segmentation tasks. The detailed network architecture is illustrated in Fig. \ref{fig:network}. Apart from the original PU-Dense structure, we incorporated transposed convolution \cite{choy20194d} to upsample features from different scales of the encoder. These features were then concatenated as input, and the SENet \cite{hu2018squeeze} is applied to learn the importance of each channel. Ultimately, a linear classifier was employed to accomplish the semantic segmentation. Besides, we adjusted the expansion factor (kernel size) for coordinate generation within the PU-Dense network to accommodate large-scale scenes.

\subsubsection{Loss Function}
For the upsampling branch, we employed a voxel-based BCE Loss consistent with PU-Dense, while for the semantic segmentation branch, we used the conventional CE Loss. The final loss for the entire network was the average of these values, with the detailed definition as follows:

\begin{equation}
  \mathcal{L}_{BCE}=-\frac{1}{N} \sum_{i}\left(x_{i} \log \left(p_{i}\right)+\left(1-x_{i}\right) \log \left(1-p_{i}\right)\right)
  \label{eq:BCE}
\end{equation}

\begin{equation}
  \mathcal{L}_{CE}=-\frac{1}{N} \sum_{i} x_{i} \log \left(p_{i}\right)
  \label{eq:CE}
\end{equation}

\begin{equation}
  \mathcal{L}=\frac{1}{2}\mathcal{L}_{BCE}+\frac{1}{2}\mathcal{L}_{CE}
  \label{eq:Loss}
\end{equation}

In $\mathcal{L}_{BCE}$, $x_i$ is the voxel label that is either occupied (1) or empty (0) in the GT point cloud. $p_i$ is the probability of the voxel being occupied and is calculated using a $sigmoid$ function applied to the decoder output $D_0$. In $\mathcal{L}_{CE}$, $x_i$ refers to the one-hot label entry, and $p_i$ represents the probability of a point being predicted as a certain class.

\subsubsection{Experiments}
To explore the relationships between multiple sub-tasks, we adopted the following experimental strategies: training solely with the upsampling branch; training solely with the semantic segmentation branch; and simultaneous training with both the upsampling and semantic segmentation branches. Such a design allowed us to quantify the phenomena of inter-task influence, providing a research basis for the design of multi-task models.

Using data with different point densities from our dataset, we took the coordinate values of the 32-channel point cloud as input, the 128-channel point cloud as the ground truth for upsampling, and the label values of the 32-channel point cloud as the ground truth for semantic segmentation. The division of the training and test sets is also done as mentioned earlier: by using a random split. Due to the sheer volume of the original data, we only use parts of Town03, 05, and 10 here. In addition, using SemanticKITTI dataset, we performed random 2x downsampling of the original 64-channel point cloud to simulate a 32-channel point cloud. We then used this data to perform the same training and directly tested the trained network on the WHU-Synthetic dataset to verify its transfer effect.

For a fair comparison, the Adam \cite{kingma2014adam} and SGD \cite{ruder2016overview} optimizer was used with a weight decay of 0.0001 and a default learning rate of 0.002. We used exponentialLR \cite{li2019exponential} to adjust the learning rate and the gamma is 0.5. During the training procedure, random rotation along the z-axis was performed for the augmentation. For the remaining hyperparameters, we strictly followed the settings used in PU-Dense. The model was trained for 10 epochs with a batch size of 4 under each setting.

\begin{table*}[th]
    \centering
    \caption{Semantic segmentation results of the multi-task experiment on the validation set. Mean IoU (mIoU [\%]), and per-class IoU [\%] scores are reported.}
    \label{tab:multi_experi}
    \renewcommand\arraystretch{1.3}
    \resizebox{\linewidth}{!}{
    \begin{tabular}{lcccccccccccccccccc}
    \toprule
    \textbf{Branch} & \textbf{mIoU $\uparrow$} & \rotatebox{90}{car} & \rotatebox{90}{bicycle} & \rotatebox{90}{motorcycle} & \rotatebox{90}{truck} & \rotatebox{90}{other-vehicle} & \rotatebox{90}{person} & \rotatebox{90}{rider} & \rotatebox{90}{road} & \rotatebox{90}{sidewalk} & \rotatebox{90}{ground} & \rotatebox{90}{building} & \rotatebox{90}{fence} & \rotatebox{90}{vegetation} & \rotatebox{90}{terrain} & \rotatebox{90}{pole} & \rotatebox{90}{traffic-sign} \\
    \midrule
    Seg. & 43.6 & 62.5 &0.0 &0.0 &60.7 &0.0 &0.0 &8.8 &93.2 &74.8 &53.1 &70.9 &67.0 &69.3 &63.3 &38.6 &35.4 \\
    Ups.+Seg. & \textbf{45.3} &64.3 &0.0 &0.0 &69.3 &0.0 &0.0 &12.3 &93.3 &75.4 &54.1&71.9 &73.7 &72.3 &63.8 &40.2 &34.7 \\
    \bottomrule
    \end{tabular}
    }
\end{table*}

\begin{table}[!t]
\centering
\caption{We conducted experiments on the \textbf{upsampling branch} (Ups.), the \textbf{semantic segmentation branch} (Seg.), and both branches simultaneously (Ups.+Seg.). Chamfer distance(CD, multiplied by 10$^3$) and average recall (AR [\%]) are reported as evaluation metrics for upsampling and mean IOU (mIoU [\%]) is reported for semantic segmentation. $^{\dag}$ We use the pretrained model obtained from SemanticKITTI to test on WHU-Synthetic.}
\label{tab:multi}
\renewcommand\arraystretch{1.3}
\resizebox{0.9\linewidth}{!}{
\begin{tabular}{lccc}
\toprule
\textbf{Branch} & \textbf{Ups. AR $\uparrow$} & \textbf{Ups. CD $\downarrow$} & \textbf{Seg. mIoU $\uparrow$} \\
\midrule
Ups. & 43.2 & 45.5 & - \\
Seg. & - & - & 43.6 \\
\textbf{Ups. + Seg.} & \textbf{46.5} & \textbf{43.1} & \textbf{45.3} \\
\midrule
SemanticKITTI & 50.9 & 40.2 & 36.3 \\
Transfer to WHU-Synthetic$^{\dag}$ & 12.3 & 190.7 & 25.2 \\
\bottomrule
\end{tabular}
}
\end{table}

The experimental results are shown in Tab. \ref{tab:multi}. From the experimental results, it could be observed that when the upsampling and segmentation branches are conducted simultaneously, both the Average Recall (AR) of the upsampling task and the mean Intersection over Union (mIoU) of the segmentation task showed improvement. They increased from 43.2\% to 46.5\% and from 43.6\% to 45.3\%, respectively. It demonstrated the phenomenon of mutual promotion among specific downstream sub-tasks. The upsampling task forced the encoder to capture the implicit surface structure of objects, and such information can provide geometric constraints for semantic segmentation. Conversely, the semantic information provided by the semantic segmentation task could also assist the upsampling task in obtaining category-related information, thereby improving the upsampling process. 

At the same time, we can also see that the results of training on SemanticKITTI are comparable to training on WHU-Synthetic, and even slightly better for up-sampling. However, when the model is transferred to WHU-Synthetic for direct testing, the results are less than optimal, due to the fact that it is not reasonable to get the training data by downsampling 64 channels of data, which is not representative of the real input situation, and thus there is a great performance degradation. Additionally, WHU-Synthetic covers multiple scenarios and provides point clouds with multiple channels, which also ensures the generalization of the model trained on the WHU- Synthetic. Based on the above advantages, the model trained on WHU-Synthetic in the previous chapter showed excellent results in the up-sampling on KITTI (Sec. V), which reflects the ability of our dataset to support 3D multitasking training and provide new horizons for applications in real scenarios.

Through experimental setups, we could preliminarily leverage the proposed dataset for multi-task learning. This demonstrated the advantage of collecting the dataset from the same scenarios, allowing for the convenient integration of multiple tasks without the need for complex mappings. It also reflected that our dataset can support the exploration of the future framework.

\subsection{Place Recognition \& Reconstruction}
The simulated spatiotemporal data provided by WHU-Synthetic can enhance the robustness of recognition. Additionally, the extra high-density point clouds and city-level mesh models offer further support for feature descriptors in the reconstruction branch.

\begin{figure}[ht]
    \centering
    \includegraphics[width=\linewidth]{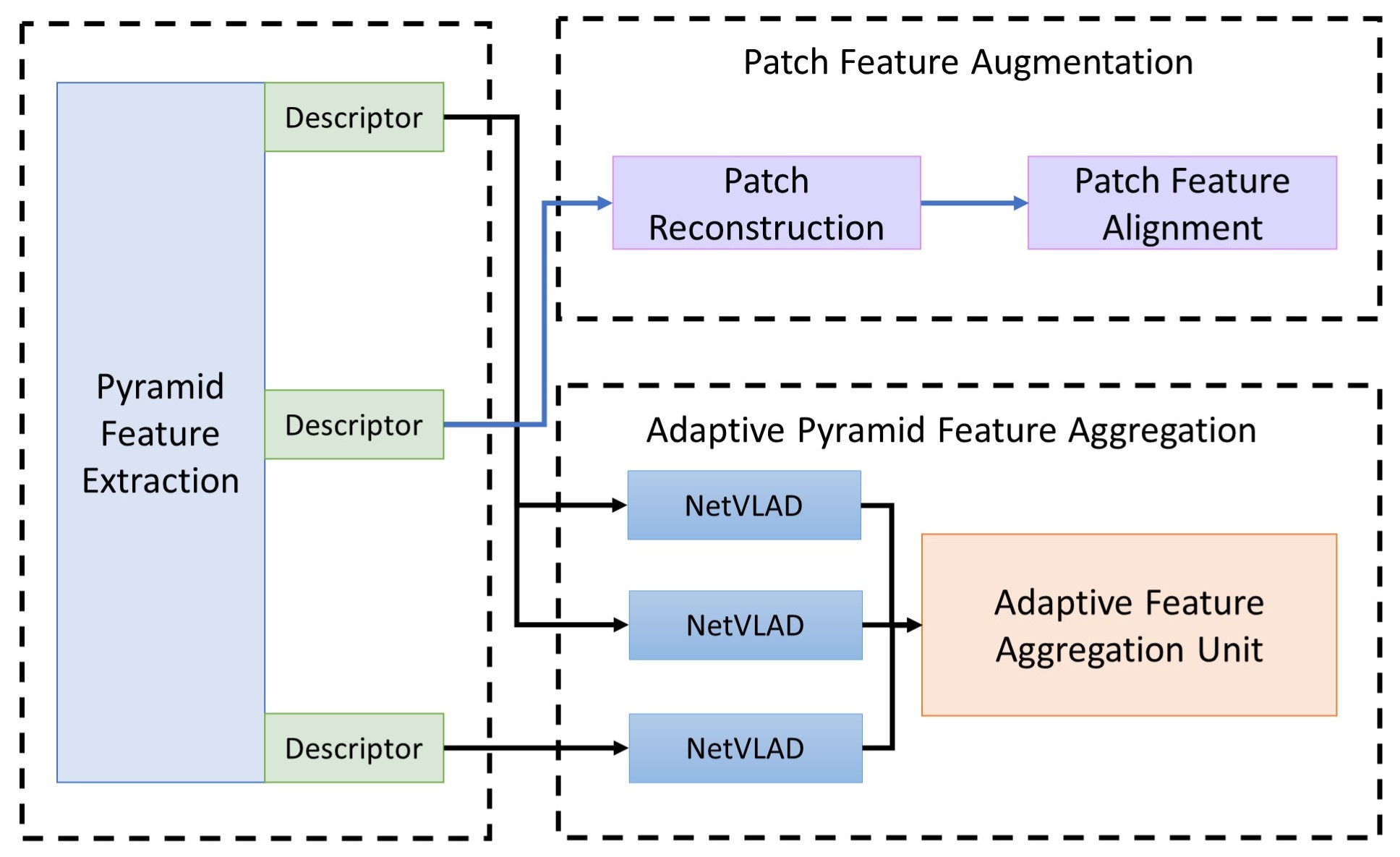}
    \caption{PatchAugNet \cite{zou2023patchaugnet} employs \textbf{patch reconstruction} to assist feature extraction.}
    \label{fig:Patch}
\end{figure}

\begin{table}[ht]
\centering
\caption{Benchmarking results of PatchAugNet \cite{zou2023patchaugnet} on place recognition. }
\label{tab:PatchAugNet}
\renewcommand\arraystretch{1.3}
\resizebox{0.57\linewidth}{!}{
\begin{tabular}{lccc}
\toprule
\textbf{Data} & \textbf{AR@1 $\uparrow$} & \textbf{AR@5 $\uparrow$} \\
\midrule
Org. & 93.5 & 95.6 \\
Org.+Mig. & 98.0 & 98.7 \\
\bottomrule
\end{tabular}
}
\end{table}

We utilized PatchAugNet \cite{zou2023patchaugnet} to explore the relationship between place recognition and reconstruction within the WHU-Synthetic dataset.

PatchAugNet demonstrated superior performance in place recognition tasks compared to other methods. As shown in Fig. \ref{fig:Patch}, The key to this improvement lies in the network's ability to utilize patch reconstruction to enhance the feature representation of the point clouds. This reconstruction process allowed PatchAugNet to capture subtle environmental changes and fine-grained details, which are crucial for accurate place recognition. The results, as shown in Tab. \ref{tab:PatchAugNet}, indicated that PatchAugNet achieved an AR@1 of 93.5\% and an AR@5 of 95.6\% when trained on the Ori. data, and an impressive AR@1 of 98.0\% and AR@5 of 98.7\% when trained on both the Ori. and Mig. data.

The test of PatchAugNet on the WHU-Synthetic dataset underscored the potential of reconstruction techniques in aiding place recognition tasks. By reconstructing patches of the environment, the network could better understand and differentiate between various locations, even when minor changes occur over time. This capability is particularly valuable in dynamic environments where conditions can vary significantly.

Furthermore, WHU-Synthetic's comprehensive and multi-task nature provides an excellent platform for exploring the interplay between reconstruction and place recognition tasks. Since the dataset includes data for multiple tasks collected within the same environmental domain, it inherently supports the investigation of how these tasks can mutually benefit each other. This alignment allowed for a deeper understanding of how reconstruction can enhance place recognition and vice versa, potentially leading to more robust and versatile 3D perception models.

\subsection{Loop Closure Detection \& Segmentation}
Using the multi-density point cloud data with semantic and instance annotations, along with the labeled ground truth point clouds obtained from surface sampling provided by WHU-Synthetic, the richness of information for loop closure detection can be enhanced, thereby improving the accuracy of reconstruction recognition.

\begin{figure}[ht]
    \centering
    \includegraphics[width=\linewidth]{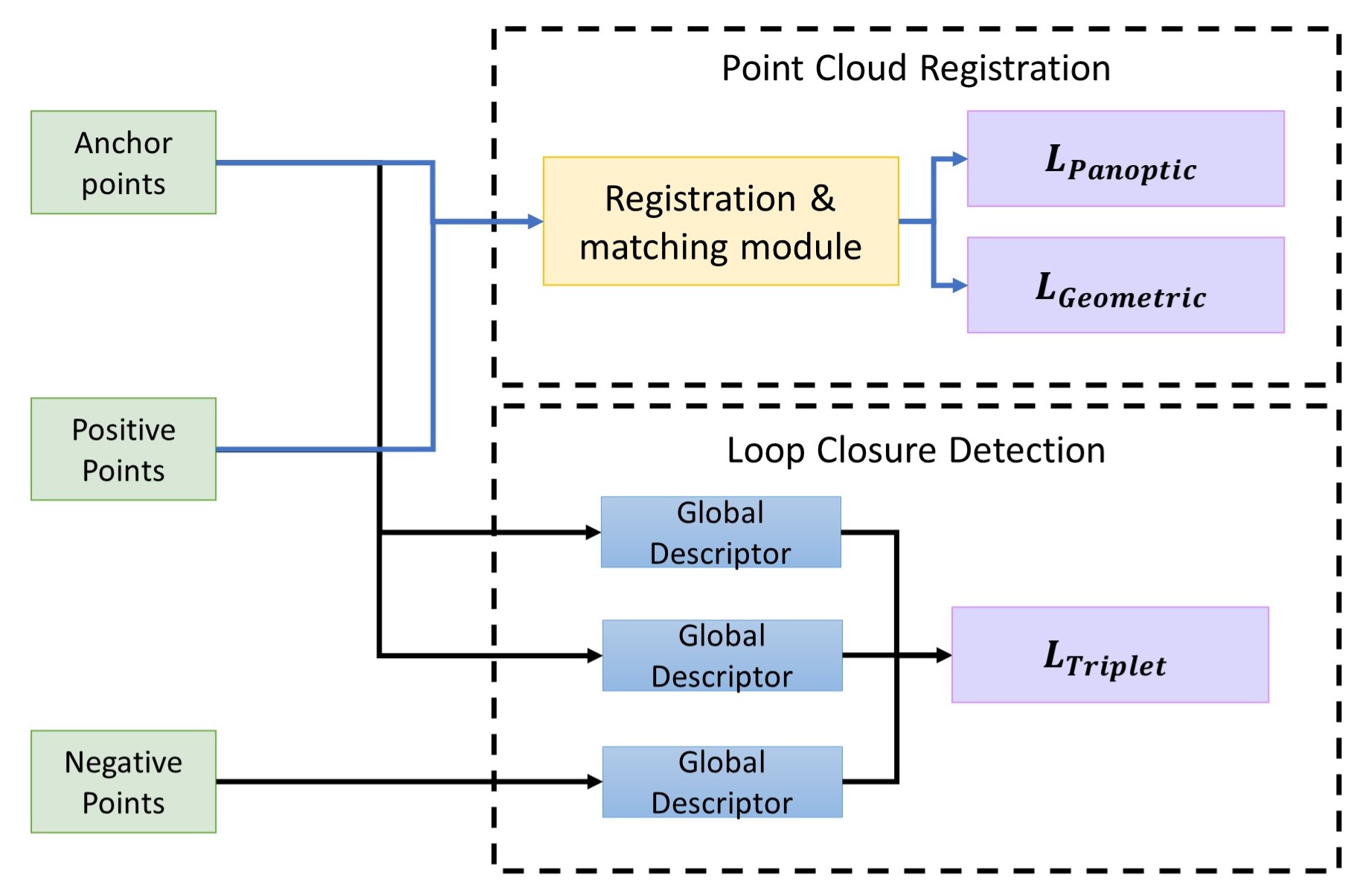}
    \caption{PADLoC \cite{arce2022padloc} uses \textbf{panoptic information} to create a branch loss in registration \& matching.}
    \label{fig:PAD}
\end{figure}

\begin{table}[ht]
\centering
\caption{Benchmarking results of PADLoC and LCDNet on loop closure detection and registration. Maximum F1 score (Max-F1 [\%]), Area Under the Curve (AUC [\%]), rotation
error ($r_{err}$ [°]) and translation error ($t_{err}$ [m]) are reported.}
\label{tab:PADLoC}
\renewcommand\arraystretch{1.3}
\resizebox{0.8\linewidth}{!}{
\begin{tabular}{lccccc}
\toprule
\textbf{Method} & \textbf{Max-F1 $\uparrow$} & \textbf{AUC $\uparrow$} & \textbf{$r_{err} \downarrow$} & \textbf{$t_{err} \downarrow$} \\
\midrule
PADLoC \cite{arce2022padloc}  & \textbf{93.6} & \textbf{87.7} & \textbf{1.21} & \textbf{1.02} \\
LCDNet \cite{cattaneo2022tro} & 88.5 & 84.0 & 1.32 & 1.08\\
\bottomrule
\end{tabular}
}
\end{table}

We utilized PADLoC \cite{arce2022padloc} to perform loop closure detection on the WHU-Synthetic dataset. As shown in Fig. \ref{fig:PAD}, this network leveraged semantic and panoptic labels to enhance the detection accuracy. The semantics of the scene played a crucial role in assisting loop closure detection by providing contextual understanding, which helps in differentiating between visually similar but contextually distinct locations.

The results, as shown in Tab. \ref{tab:PADLoC}, indicated that PADLoC achieved a Max-F1 score of 93.6 and an AUC of 87.7, outperforming LCDNet, which achieved a Max-F1 score of 88.5 and an AUC of 84.0. This demonstrated that PADLoC's use of semantic and panoptic labels allows for better feature extraction, leading to more accurate loop closure detection. The semantic information helped in identifying unique features and patterns that remain consistent over time, thereby reducing the drift accumulated in SLAM systems.

Overall, the experimental results underscored the importance of semantics in enhancing loop closure detection and demonstrated that the WHU-Synthetic dataset is well-suited for investigating the synergistic relationship between multiple 3D perception tasks. The integration of semantic information in loop closure detection highlighted the potential for improving the effectiveness of SLAM systems.

\section{Discussion and Conclusion}
We introduce WHU-Synthetic, a synthetic perception dataset featuring the exploration of 3D multi-task models, including depth completion, segmentation, upsampling, place recognition, and 3D reconstruction. We also implement innovative sensor settings, endowing our dataset with unique characteristics. For instance, it can provide data with varying densities within the same frame, offer surface-sampled data at a city level, and simulate environmental changes. These features aim to contribute to the advancement of the field of point cloud perception. In addition, benefiting from the carefully designed settings in our data collection, we integrate the upsampling and semantic segmentation task into a multi-task framework. Through multi-task experiments, we demonstrate the mutually beneficial relationship among different sub-tasks.

However, we also acknowledge the limitations of our dataset. Joint training with only two subtasks can only provide preliminary evidence of the relevance of certain 3D perception tasks. To achieve a truly multi-task network, further effective experiments and explorations are needed. We plan to progressively incorporate more sub-tasks in the future to continue our research.

Additionally, although virtual environments possess unique advantages, one might argue that the domain gap between synthetic and real data is a weakness. In our defense, we believe that when synthetic data is used correctly, it can be used to enhance task performance on real data. For example, \cite{deschaud2021paris} indicates that the dataset generated by the CARLA simulator can improve the accuracy of point cloud semantic segmentation to a certain extent. \cite{Sallab_Sobh_Zahran_Shawky_2019} showed that augmenting with point clouds generated from the CARLA simulator can improve bird's eye view 2D detection performance on the KITTI dataset. \cite{xiao2022transfer, hu2020towards, espadinha2021lidar} developed several domain adaptation methods to mitigate domain shift on synthetic point cloud data as much as possible. Furthermore, we also adjusted the sensor settings and added noise on LiDAR to minimize domain shift as much as possible. At the same time, our simulation of temporal changes was limited to short-term variations. Simulating longer-term changes, such as seasonal shifts, would require significantly more effort and resources to fully capture these phenomena. Such an in-depth simulation would be necessary to thoroughly investigate and quantify the impact of these important temporal features on place recognition tasks.

In summary, by introducing WHU-Synthetic, we hope that our dataset can aid researchers in conducting more novel preliminary explorations under the current scarcity of 3D datasets, to better realize more grounded concepts in the future. We believe our work is a valuable and necessary first step towards 3D multi-task learning and deserves sharing with the community.

\bibliographystyle{IEEEtran}
\bibliography{main}
\begin{IEEEbiography}[{\includegraphics[width=1in,height=1.25in,clip,keepaspectratio]{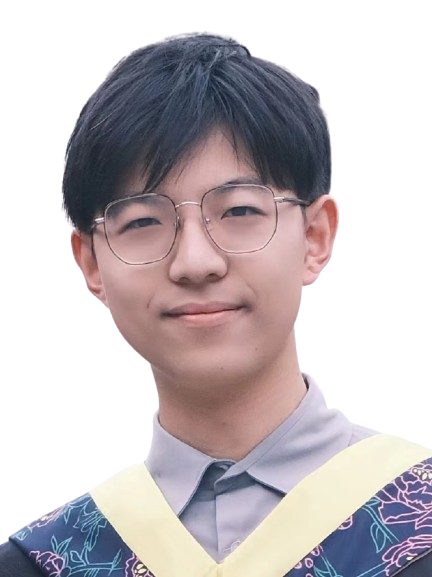}}]{Jiahao Zhou} received the B.E. degree in Remote Sensing from Wuhan University, Wuhan, China in 2024. He is currently pursuing the M.E. degree in Safety Engineering at Tsinghua University, Beijing, China.

His research interests include 3-D multi-task learning, point cloud processing and disaster situational awareness.
\end{IEEEbiography}

\begin{IEEEbiography}[{\includegraphics[width=1in,height=1.25in,clip,keepaspectratio]{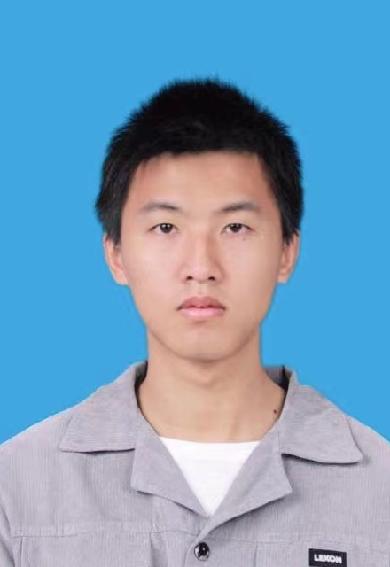}}]{Chen Long} received the B.S. degree from the School of Geodesy and Geomatics, Wuhan University, Wuhan, China, in 2021. He is currently pursuing a doctoral degree in the State Key Laboratory of Information Engineering in Surveying, Mapping, and Remote Sensing.

His research interests include enhancing point cloud quality and 3-D shape estimation.
\end{IEEEbiography}

\begin{IEEEbiography}[{\includegraphics[width=1in,height=1.25in,clip,keepaspectratio]{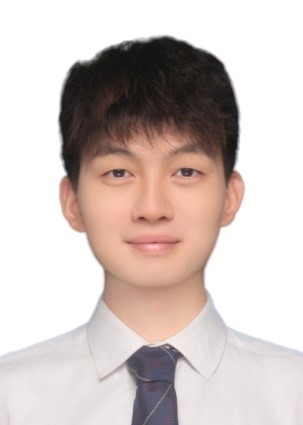}}]{Yue Xie} received the B.S., and M.S. degrees in remote sensing from Wuhan University, China, in 2021 and 2024, respectively. His current research interests include localization and robotic perception.

His research interests include enhancing point cloud quality and 3-D shape estimation.
\end{IEEEbiography}

\begin{IEEEbiography}[{\includegraphics[width=1in,height=1.25in,clip,keepaspectratio]{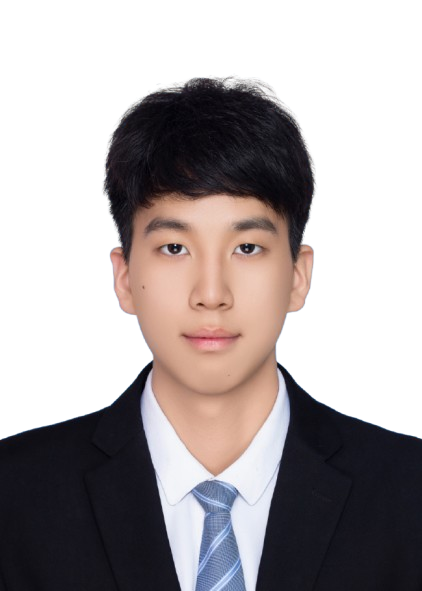}}]{Jialiang Wang} received his B.E. degree from Wuhan University, Wuhan, China, in 2024. He is currently pursuing the Ph.D. degree in mechanical and automation engineering at The Chinese University of Hong Kong, Shatin, Hong Kong. 

His research interests include SLAM, intelligent perception and unmanned systems.
\end{IEEEbiography}

\begin{IEEEbiography}[{\includegraphics[width=1in,height=1.25in,clip,keepaspectratio]{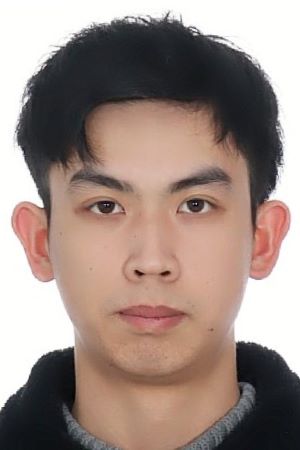}}]{Conglang Zhang} received the B.S. degree from the School of Geodesy and Geomatics, Wuhan University, Wuhan, China, in 2024. He is currently pursuing a master's degree in the State Key Laboratory of Information Engineering in Surveying, Mapping, and Remote Sensing.

His research interests include enhancing point cloud quality and 3-D shape estimation.
\end{IEEEbiography}

\begin{IEEEbiography}[{\includegraphics[width=1in,height=1.25in,clip,keepaspectratio]{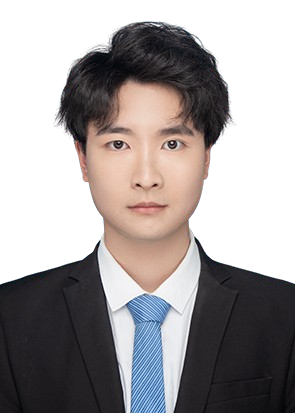}}]{Boheng Li} received his B.E. degree in Information Security from Wuhan University, China in 2024. He is currently pursuing his Ph.D. degree with the College of Computing and Data Science, Nanyang Technological University, Singapore. 

His research interests include trustworthy machine learning and AI security.
\end{IEEEbiography}

\begin{IEEEbiography}[{\includegraphics[width=1in,height=1.25in,clip,keepaspectratio]{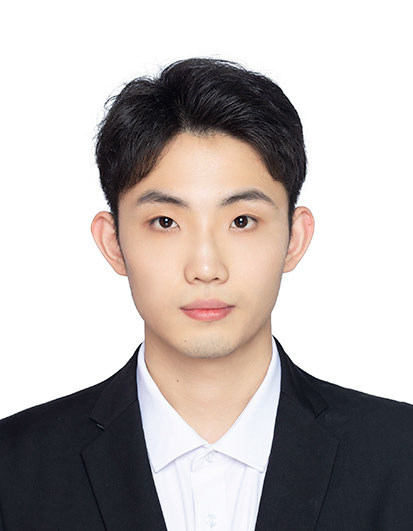}}]{Haiping Wang} is currently pursuing the Ph.D. degree in State Key Laboratory of Information Engineering in Surveying, Mapping and Remote Sensing (LIESMARS), Wuhan University, Wuhan, China, advised by Prof. Zhen Dong and Prof. Bisheng Yang. 

His research interests include the field of 3-D Computer Vision, particularly point cloud registration and 3-D reconstruction.
\end{IEEEbiography}

\begin{IEEEbiography}[{\includegraphics[width=1in,height=1.25in,clip,keepaspectratio]{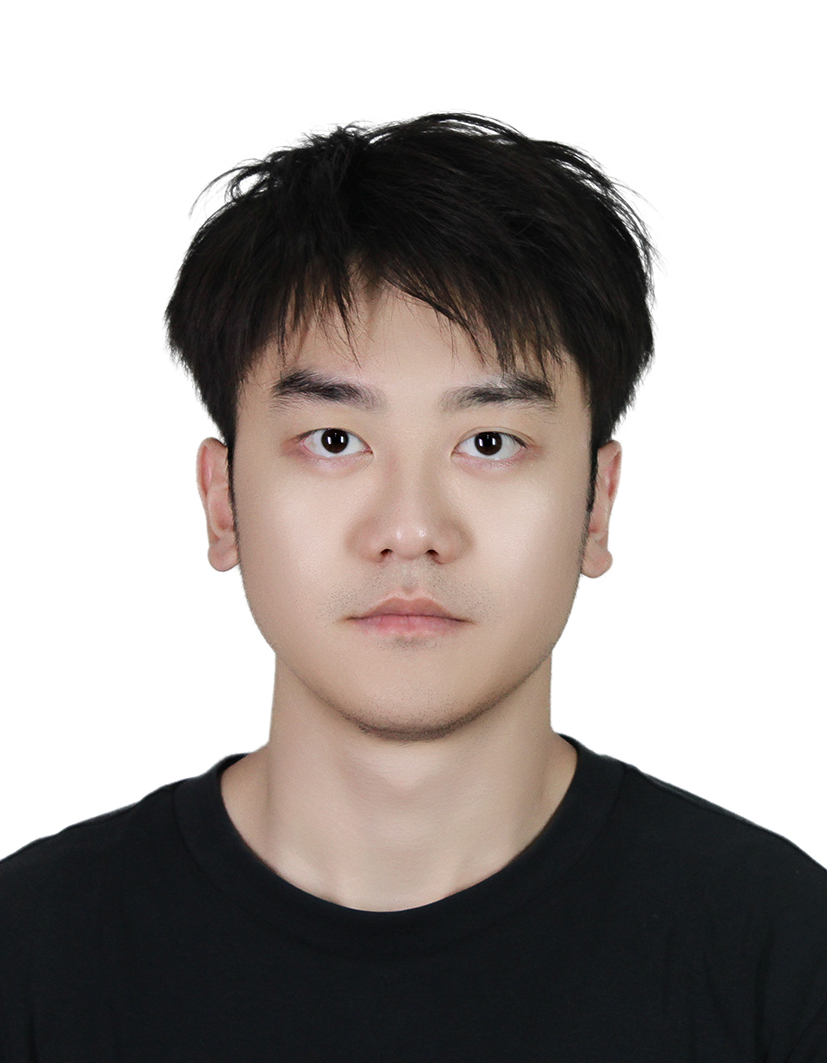}}]{Zhe Chen} received the B.S. degree from the School of Earth Sciences and Engineering, Hohai University, Nanjing, China, in 2020. He is currently pursuing the Ph.D degree in the State Key Laboratory of Information Engineering in Surveying, Mapping, and Remote Sensing, Wuhan University, Wuhan, China.

His research interests include image/point cloud processing and their applications in urban morphology.
\end{IEEEbiography}

\begin{IEEEbiography}[{\includegraphics[width=1in,height=1.25in,clip,keepaspectratio]{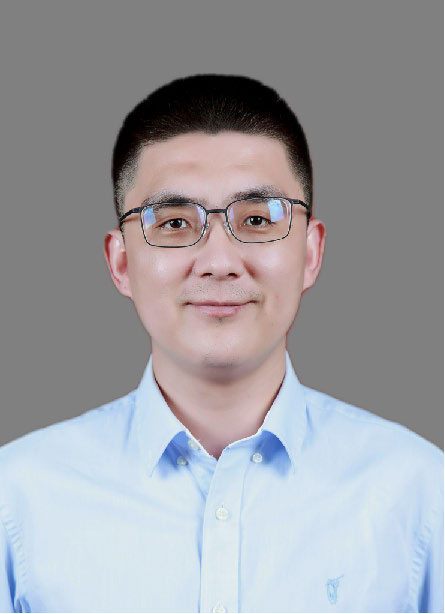}}]{Zhen Dong} (Member, IEEE) received the B.E. and Ph.D. degrees in remote sensing and photogrammetry from Wuhan University, Wuhan, China, in 2011 and 2018, respectively.

He is currently a Professor with the State Key Laboratory of Information Engineering in Surveying, Mapping, and Remote Sensing (LIESMARS), Wuhan University. His research interests include the field of 3-D computer vision, particularly 3-D reconstruction, scene understanding, and point cloud processing as well as their applications in intelligent transportation systems, digital twin cities, urban sustainable development, and robotics.
\end{IEEEbiography}

\newpage

\begin{figure*}[ht]
	\centering
	\subfloat[]{\includegraphics[width=0.32\linewidth]{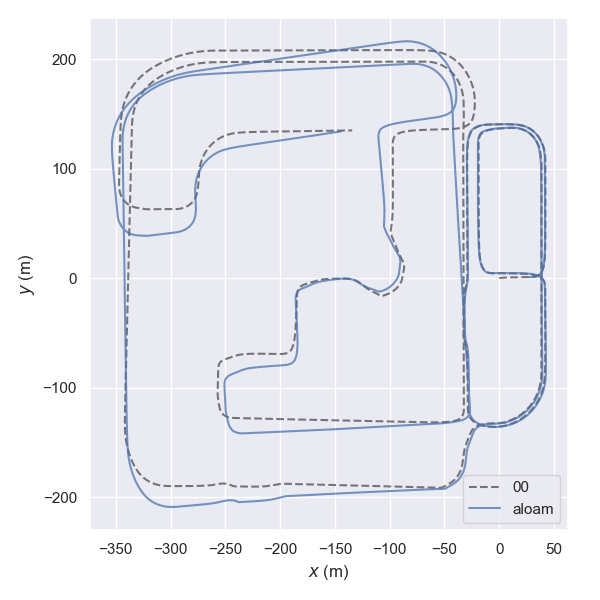}}
    \hfil
	\subfloat[]{\includegraphics[width=0.32\linewidth]{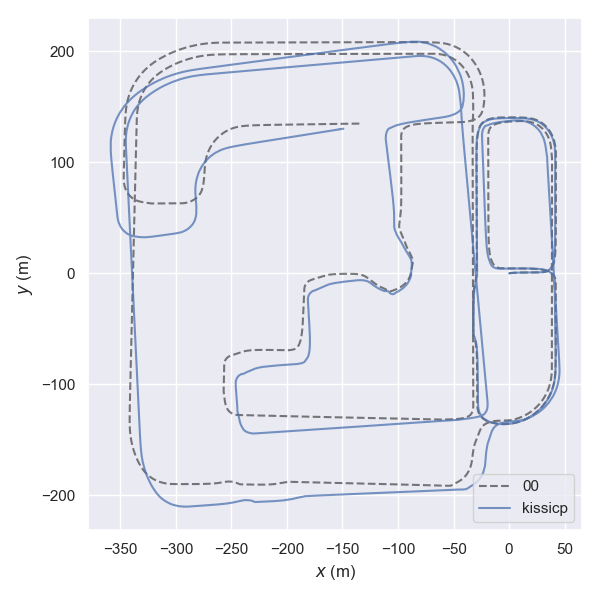}}
    \hfil
	\subfloat[]{\includegraphics[width=0.32\linewidth]{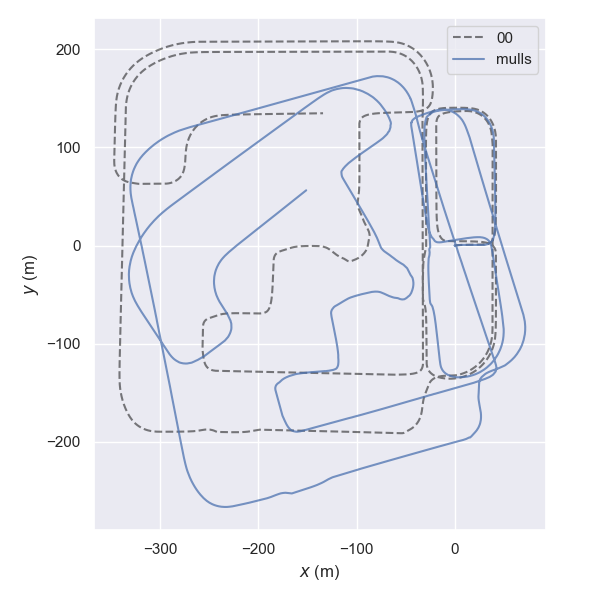}}
    \\
	\subfloat[]{\includegraphics[width=0.32\linewidth]{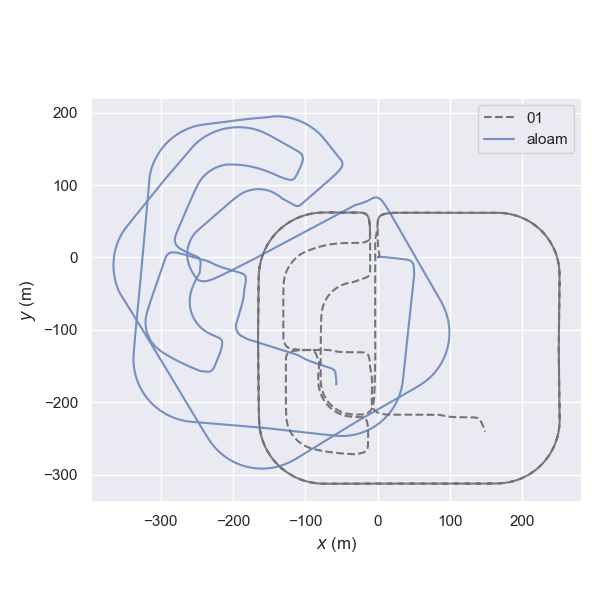}}
    \hfil
	\subfloat[]{\includegraphics[width=0.32\linewidth]{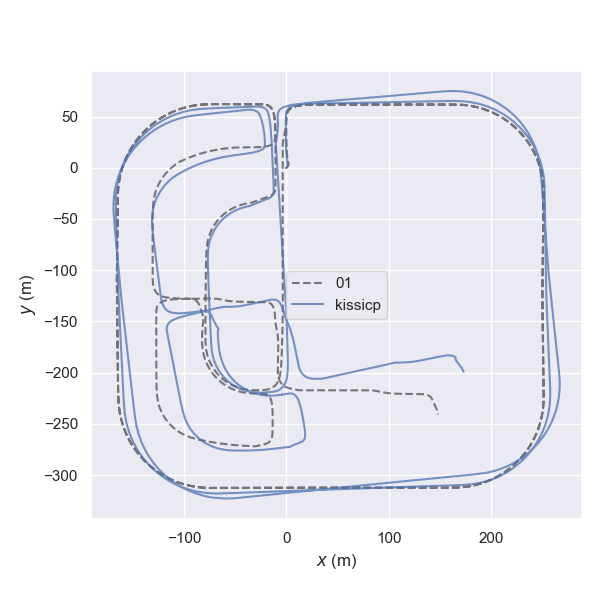}}
    \hfil
	\subfloat[]{\includegraphics[width=0.32\linewidth]{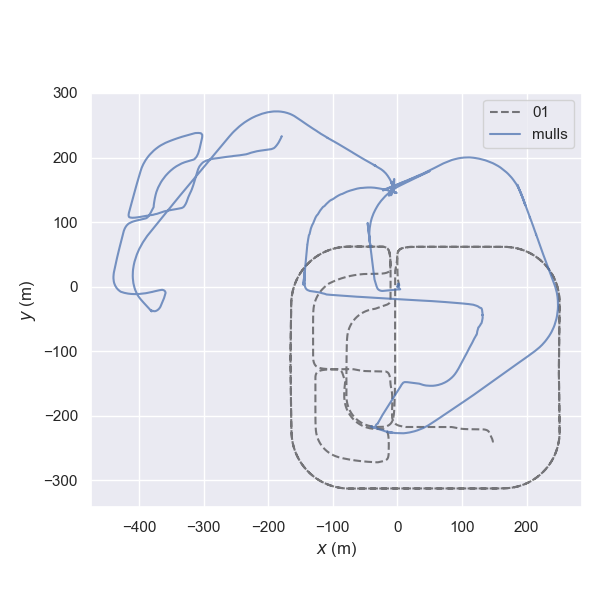}}
    \\
	\subfloat[]{\includegraphics[width=0.32\linewidth]{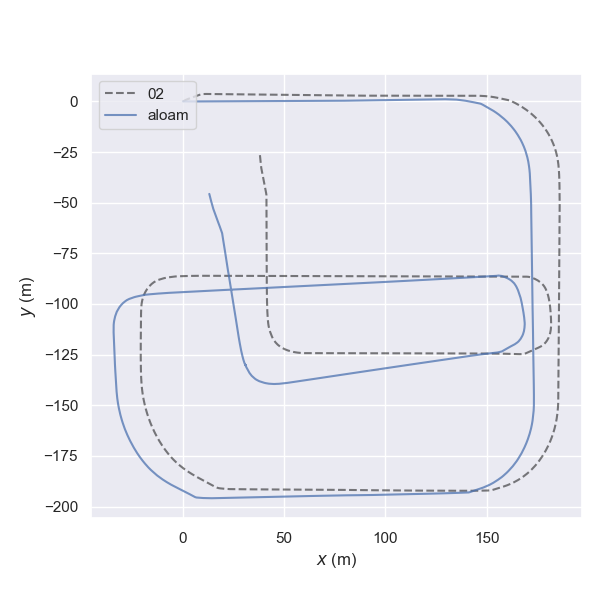}}
    \hfil
	\subfloat[]{\includegraphics[width=0.32\linewidth]{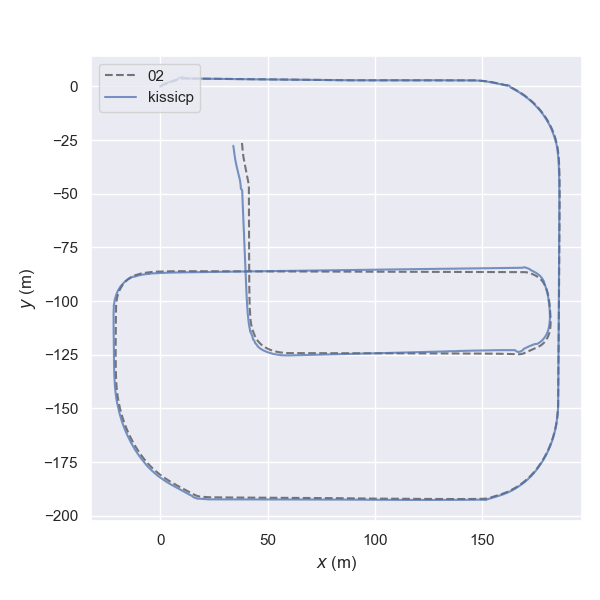}}
    \hfil
	\subfloat[]{\includegraphics[width=0.32\linewidth]{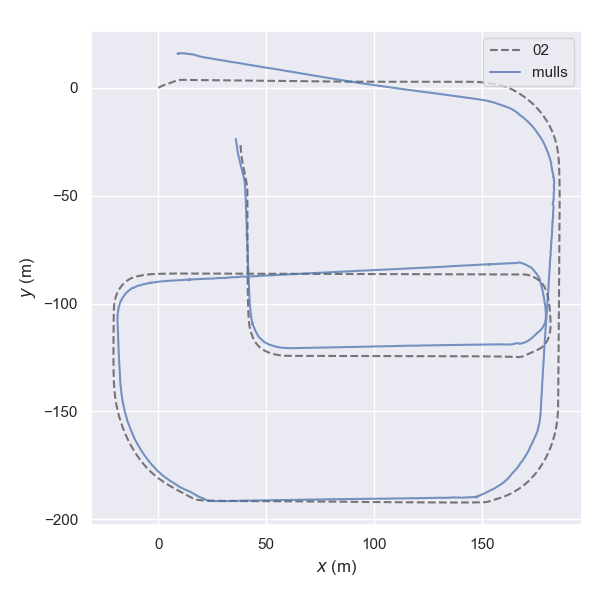}}
\caption{Visualization results of LiDAR odometry estimation on each sequence. Note that 00 refers to sequence Town03, 01 refers to sequence Town05 and 02 refers to sequence Town10. }
\label{fig:odom}
\end{figure*}

\begin{figure*}[ht]
\centering
	\subfloat[]{\includegraphics[width=0.24\linewidth]{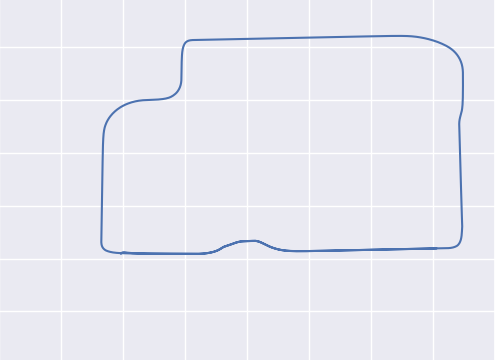}}
    \hfil
	\subfloat[]{\includegraphics[width=0.24\linewidth]{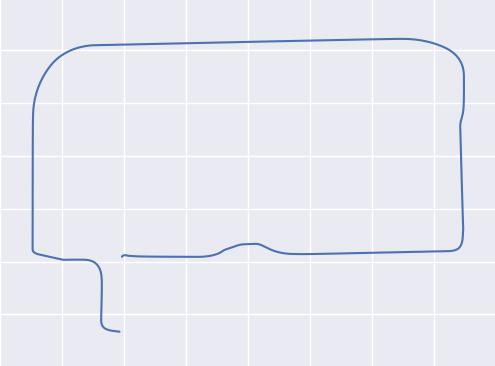}}
    \hfil
	\subfloat[]{\includegraphics[width=0.24\linewidth]{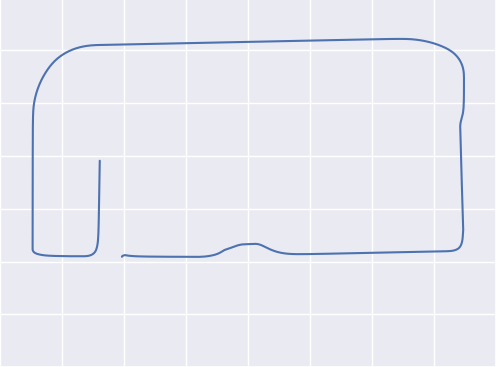}}
    \hfil
	\subfloat[]{\includegraphics[width=0.24\linewidth]{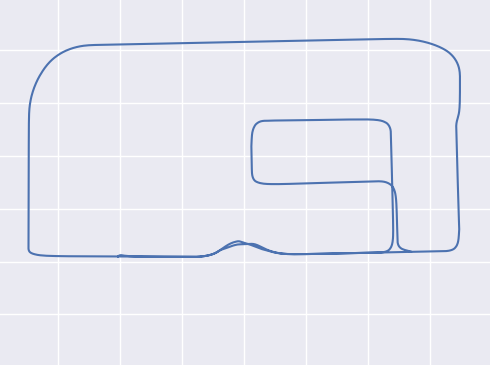}}
\caption{Visualization results of the designed 4 routes in the place recognition task. }
\label{fig:recog}
\end{figure*}

\begin{figure*}[ht]
    \centering
    \includegraphics[width=\linewidth]{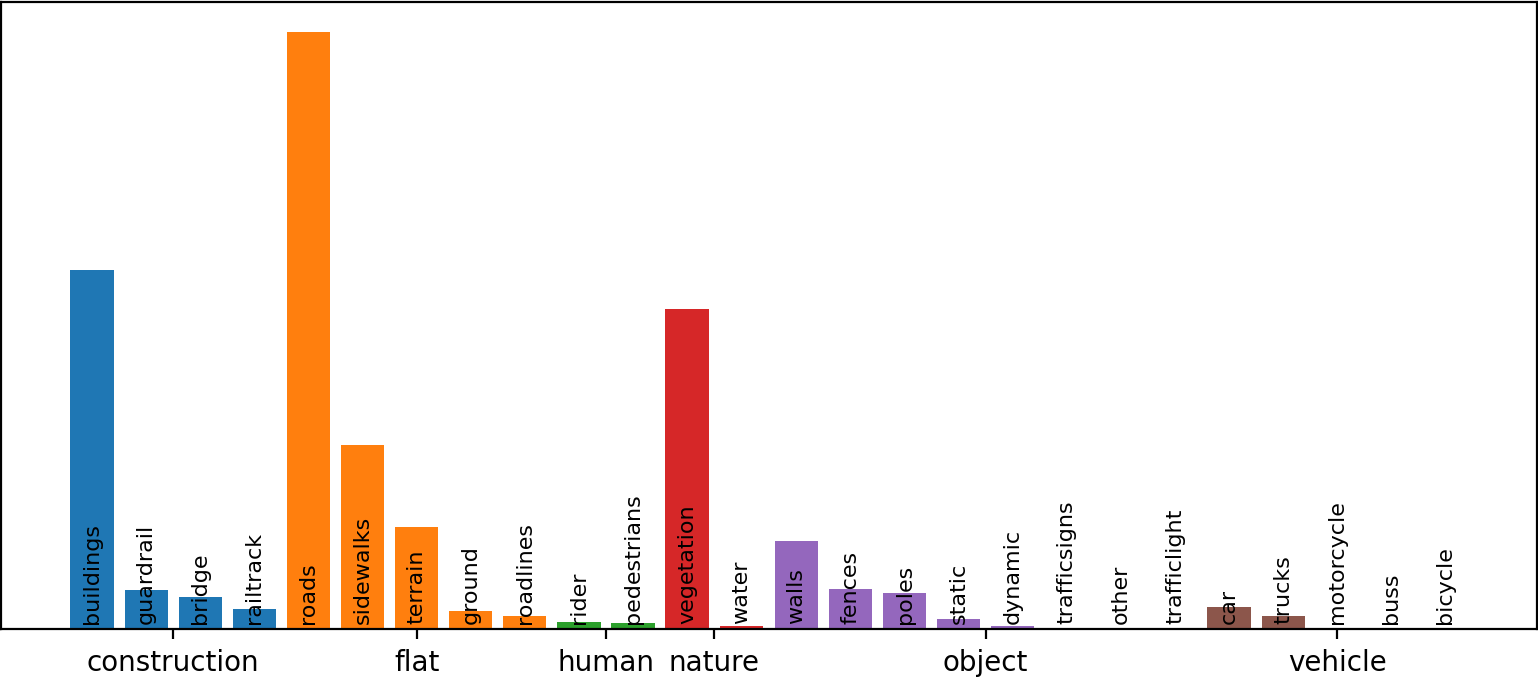}
    \caption{Class distribution in the semantic segmentation task.}
\end{figure*}

\end{document}